\documentclass{article}

 \usepackage[preprint]{neurips_2026}

\usepackage[utf8]{inputenc} 
\usepackage[T1]{fontenc}    
\usepackage{hyperref}       
\usepackage{url}            
\usepackage{booktabs}       
\usepackage{amsfonts}       
\usepackage{nicefrac}       
\usepackage{microtype}      
\usepackage{xcolor}         
\usepackage{graphicx} 
\usepackage{amsfonts}
\usepackage{amsmath}
\usepackage{amssymb}
\usepackage{colortbl}
\usepackage{array}
\usepackage{wrapfig}
\usepackage{tabularx}
\usepackage{makecell}
\usepackage{multirow}
\usepackage{subcaption}
\usepackage{cleveref}
\usepackage{tocloft}
\usepackage{arydshln}

\newcommand\blfootnote[1]{%
  \begingroup
  \renewcommand\thefootnote{}\footnote{#1}%
  \addtocounter{footnote}{-1}%
  \endgroup
}

\definecolor{fst}{HTML}{F4A6B5}   
\definecolor{sed}{HTML}{F9CFD7}  
\definecolor{thd}{HTML}{FCE6EA}   
\definecolor{default}{HTML}{D5CEE8}  

\newcommand{\fst}{\cellcolor{fst}}
\newcommand{\sed}{\cellcolor{sed}}
\newcommand{\thd}{\cellcolor{thd}}

\makeatletter
\newcommand{\TOCdisable}{%
  \let\saved@addcontentsline\addcontentsline
  \renewcommand{\addcontentsline}[3]{}%
}
\newcommand{\TOCenable}{%
  \let\addcontentsline\saved@addcontentsline
}
\makeatother

\definecolor{citecolor}{HTML}{FF758F}
\definecolor{linkcolor}{HTML}{F54747}
\hypersetup{colorlinks=true,citecolor=citecolor, linkcolor=linkcolor, urlcolor=linkcolor}

\newcounter{todos}
\AtEndDocument{\ifnum\value{todos}>0 \PackageWarning{TODOS}{There are \arabic{todos} todos left in this paper! Fix them before submitting the paper!} \fi}

\title{RAWild: Sensor-Agnostic RAW Object Detection via Physics-Guided Curve and Grid Modeling}

\author{%
  \textbf{Shuhong Liu$^{1,2,\spadesuit}$,\;
  Gengjia Chang$^{2,\spadesuit}$,\; Jun Liu$^{2}$,\; Xuangeng Chu$^{1,2}$}\\ 
  \textbf{Yinqiang Zheng$^{1}$,\; Tatsuya Harada$^{1,3}$,\; Ziteng Cui$^{1,2,\diamondsuit}$} \\
  [8pt]
  $^{1}$The University of Tokyo \quad
  $^{2}$I2WM \quad
  $^{3}$RIKEN
  \vspace{-1.5em}
}

\begin{document}

\maketitle
\blfootnote{$^\spadesuit$These authors contribute equally to this work. $^{\diamondsuit}$Corresponding author.}

\begin{figure}[!h]
  \centering
  \includegraphics[width=\linewidth]{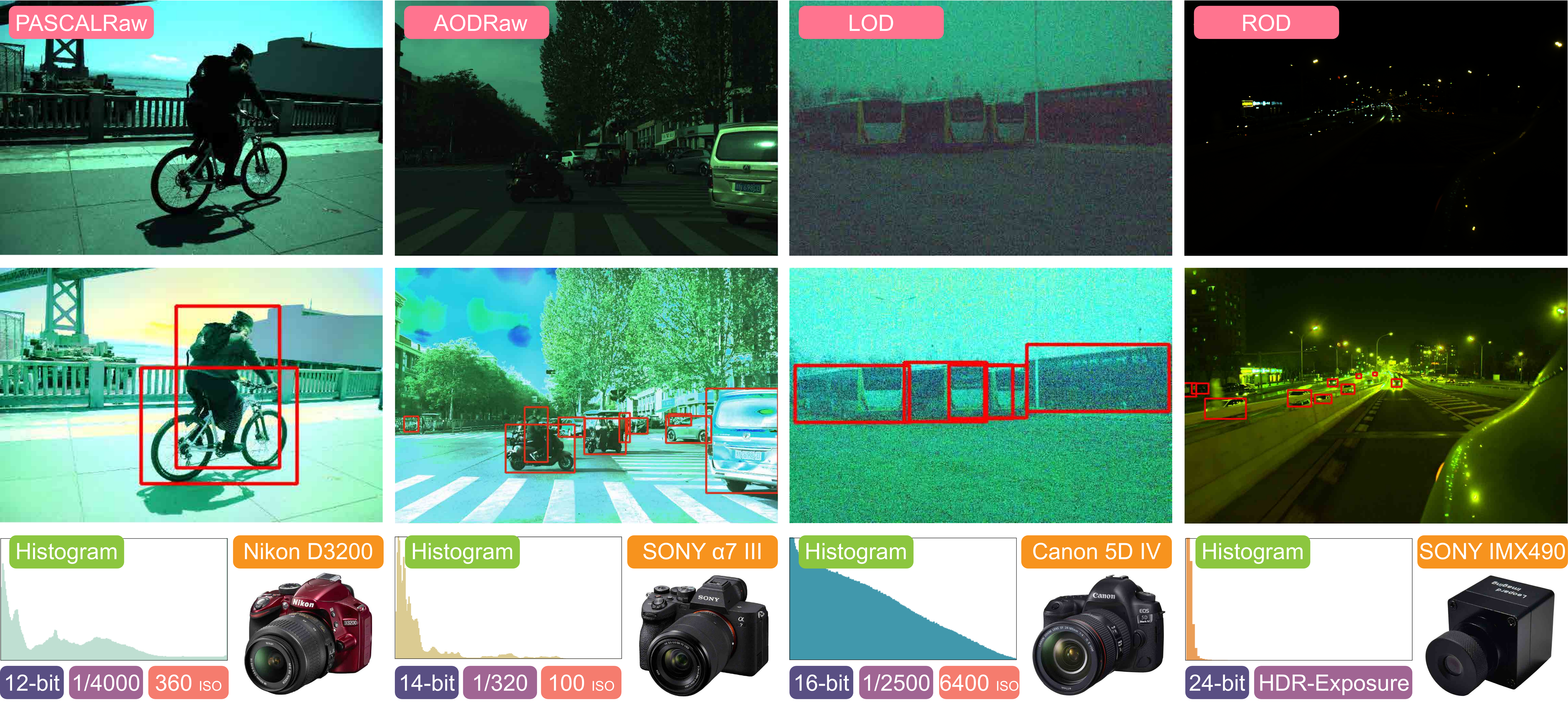}
  \caption{RAW object detection across sensors. Four benchmarks~\cite{omid2017toward,hong2021crafting,li2025towards,xu2023toward} captured by distinct sensors under different environments, exposure settings, spectral sensitivities, and bit-depths exhibit substantially different pixel distributions, posing a fundamental challenge for sensor-agnostic generalization.}
  \label{fig:teaser}
\end{figure}

\begin{abstract}
Camera sensor RAW data offers intrinsic advantages for object detection, including deeper bit depth, preserved physical information, and freedom from image signal processor (ISP) distortions. However, varying exposure conditions, spectral sensitivities, and bit depths across devices introduce substantially larger domain gaps than sRGB, making sensor-agnostic generalization a fundamental challenge. In this study, we present \textbf{RAWild}, a physics-guided global-local tone mapping framework for sensor-agnostic RAW object detection. By factoring sensor-induced variations into a global tonal correction and a spatially adaptive local color adjustment, both driven by RAW distribution priors, our framework enables a single network to train jointly across heterogeneous sensors. To further support cross-sensor generalization, we construct a physics-based RAW simulation pipeline that synthesizes realistic sensor outputs spanning diverse spectral sensitivities, illuminants, and sensor non-idealities. Extensive experiments across multiple RAW benchmarks covering bit depths from 10 to 24 demonstrate state-of-the-art (SOTA) performance under single-dataset, mixed-dataset, and challenging robustness settings.
\end{abstract}

\begin{figure}[!tp]
    \centering
    \begin{subfigure}{0.30\linewidth}
        \centering
        \includegraphics[width=\linewidth]{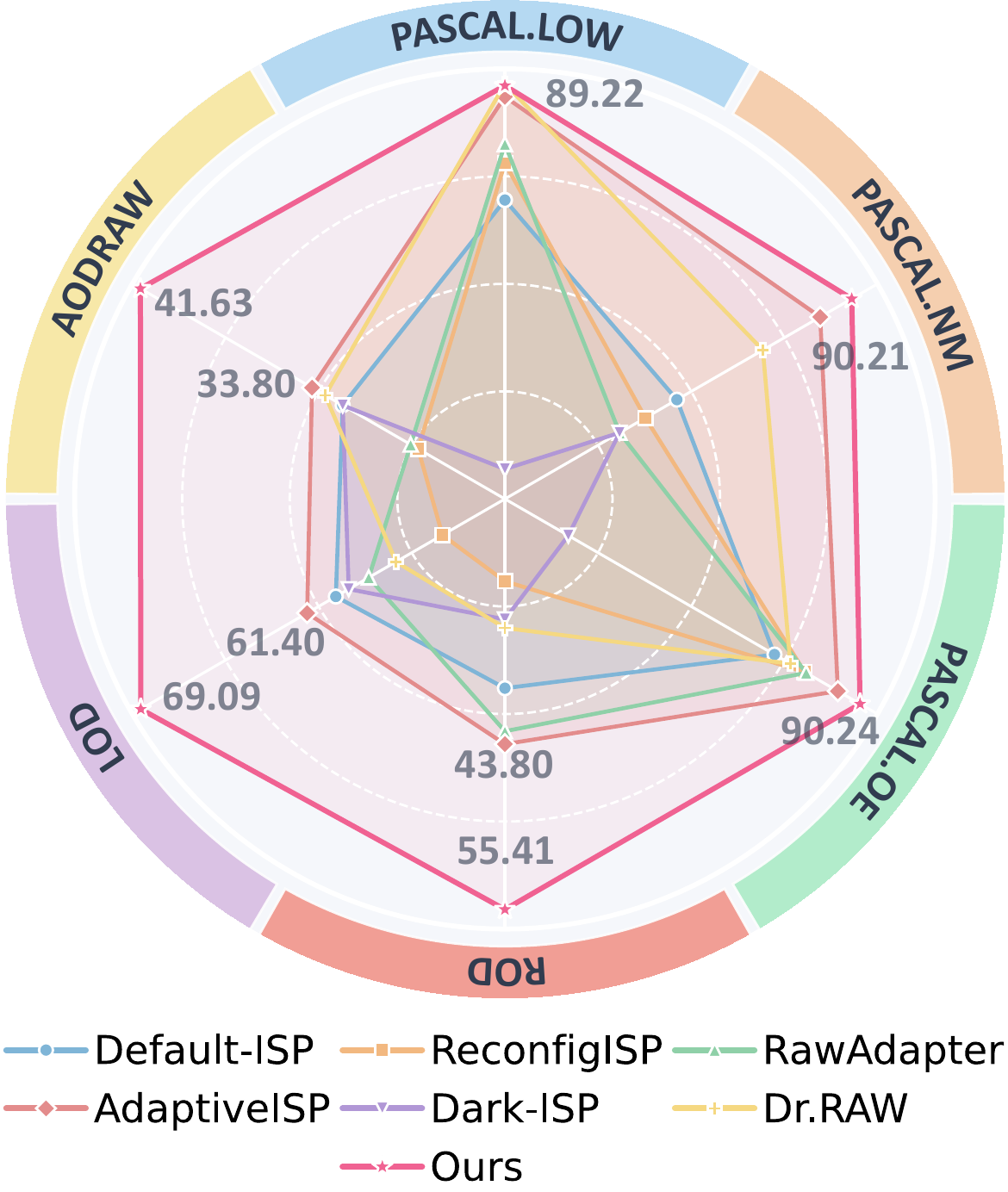}
        \vspace{-1.5em}
        \caption{Detection}
    \end{subfigure}
    \hfill
    \begin{subfigure}{0.30\linewidth}
        \centering
        \includegraphics[width=\linewidth]{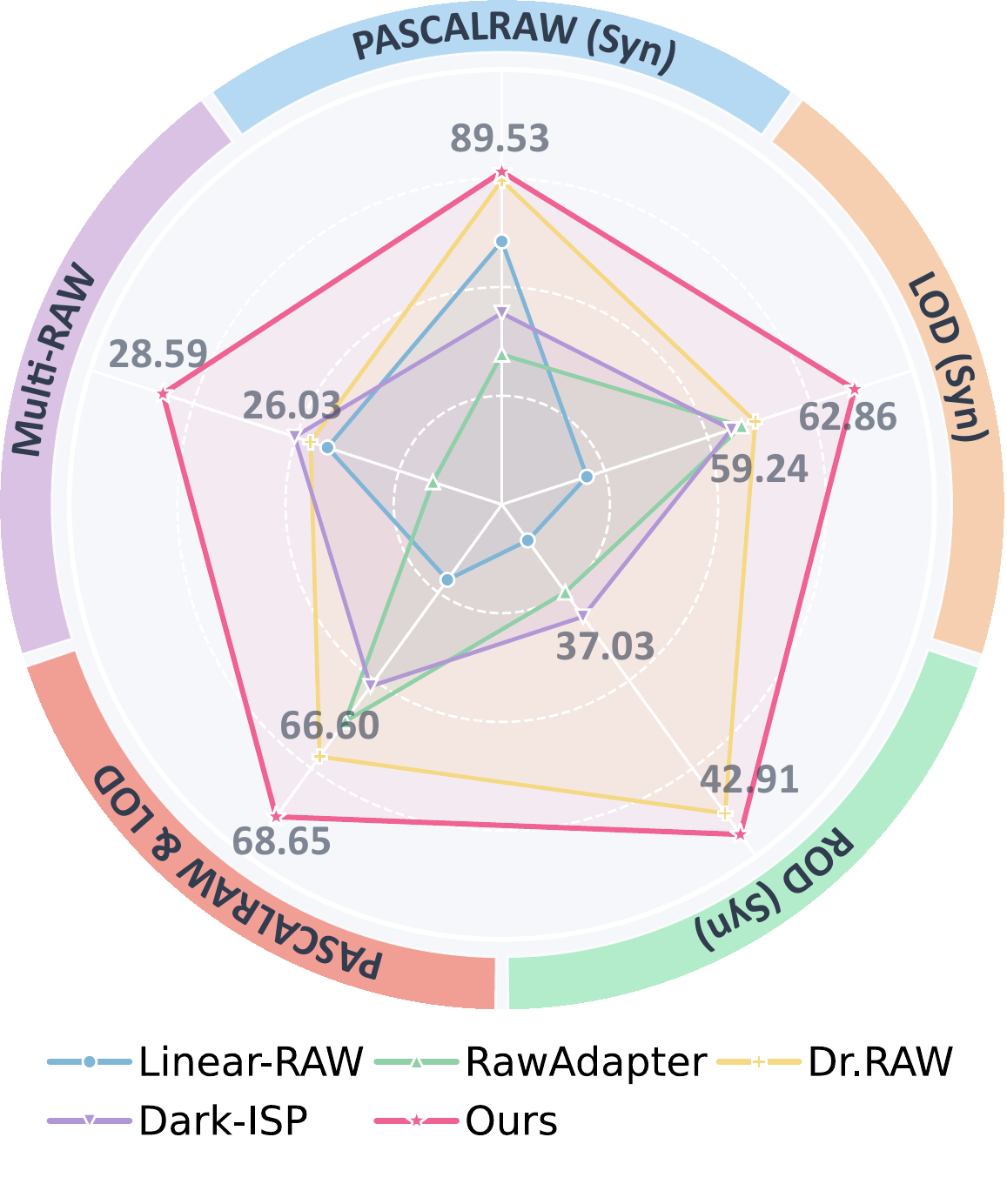}
        \vspace{-1.5em}
        \caption{Multi-dataset Detection}
    \end{subfigure}
    \hfill
    \begin{subfigure}{0.30\linewidth}
        \centering
        \includegraphics[width=\linewidth]{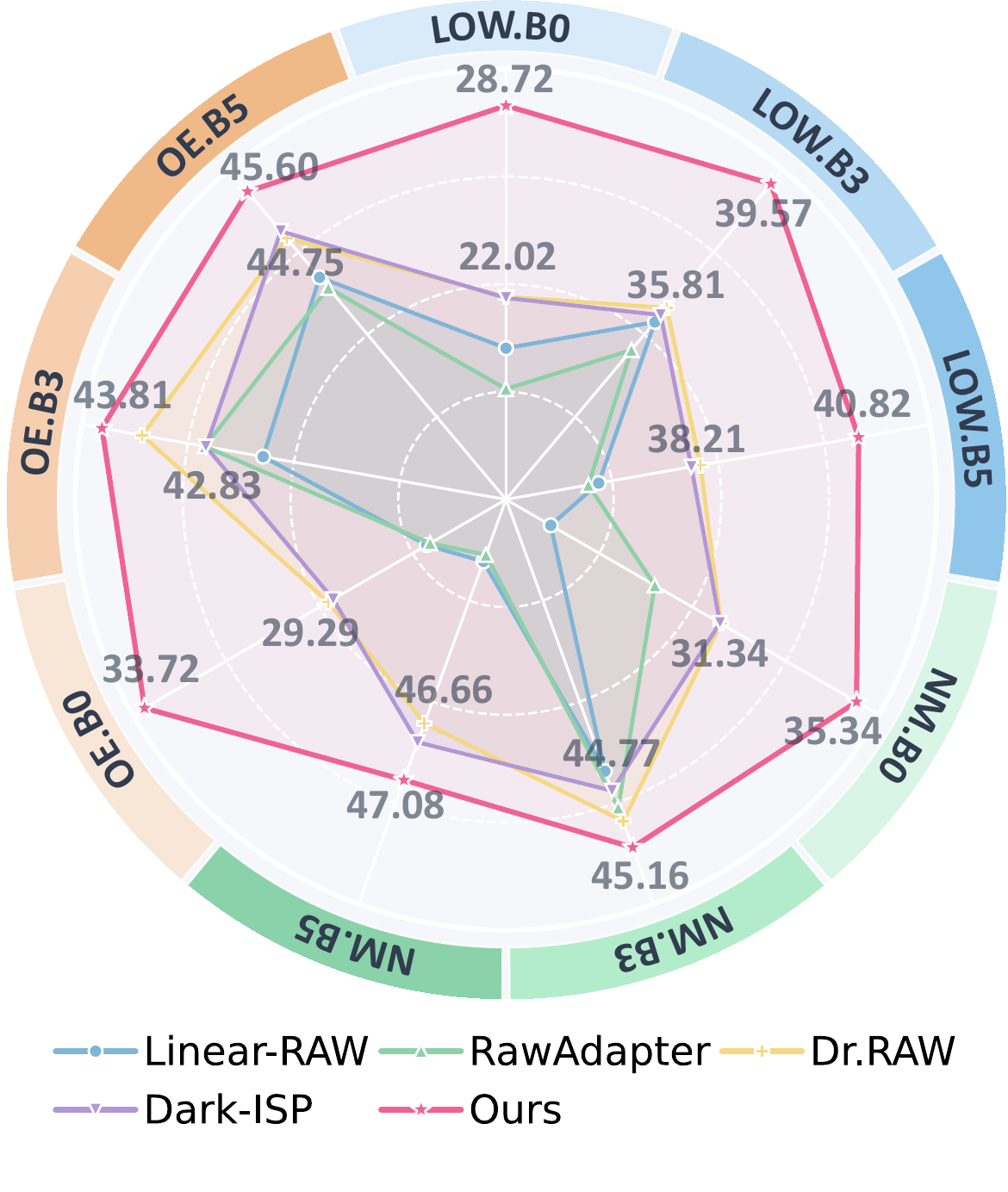}
        \vspace{-1.5em}
        \caption{Segmentation}
    \end{subfigure}
    \caption{Radar comparison with representative baselines on (a) single-dataset RAW object detection, (b) mixed-sensor dataset RAW object detection, and (c) RAW image segmentation.}
    \label{fig:radar}
\end{figure}

\section{Introduction}
\label{sec:intro}

Object detection stands as one of the most fundamental challenges in computer vision. In recent years, there has been a growing shift in focus from conventional sRGB-based detection to operating directly on camera sensor RAW data~\cite{cui2024raw,wang2024adaptiveisp,guo2025dark,xie2026simrod}, driven by several intrinsic advantages of camera RAW: deeper bit depth, richer physical information, and freedom from the nonlinear distortions introduced by the Image Signal Processor (ISP) pipeline~\cite{karaimer2016software}. These properties endow RAW-based object detection with superior robustness under extreme conditions, such as challenging illumination and adverse weather~\cite{li2025towards}, thereby holding significant practical value for real-world applications, including embodied intelligence, surveillance, and autonomous driving~\cite{saneryee2023teslafsd}.

Unlike sRGB images, which ISP pipelines normalize into a unified color space~\cite{susstrunk1999standard}, RAW sensor data preserves device-specific characteristics such as spectral sensitivity, native bit depth, and CFA layout. These properties vary considerably across cameras. For instance, \Cref{fig:teaser} shows that the Sony IMX490 sensor in ROD dataset~\cite{xu2023toward} captures 24-bit RAW images with a standard Bayer CFA, whereas the Canon 5D-IV camera in LOD dataset~\cite{hong2021crafting} records 16-bit RAW images with a distinct spectral response. Even for identical scenes, hardware discrepancies induce much larger domain gaps in RAW than in sRGB~\cite{perevozchikov2024rawformer}. More fundamentally, RAW appearance is shaped by a chain of factors spanning illumination, capture settings, and sensor hardware. The illuminant spectrum and radiance determine the incident photon flux, while shutter time and ISO gain scale the accumulated signal. The color filter response, lens-shading profile, near-saturation non-linearity, and ADC bit depth then jointly govern how the signal is recorded. Each factor varies by orders of magnitude across cameras and capture conditions, and their effects compound nonlinearly in the RAW image.

To address these discrepancies, previous methods either learn the end-to-end ISP pipeline~\cite{yu2021reconfigisp,wang2024adaptiveisp,guo2025dark} or insert learnable ISP-style blocks by stages~\cite{cui2024raw,gamrian2025beyond}. Such designs remain tightly coupled to specific sensor configurations and rely on handcrafted operator chains. Consequently, they struggle to establish a unified pipeline across heterogeneous data~\cite{radosavovic2020designing}, limiting their scalability across diverse environmental illuminations, camera responses, and bit-depths.

\textbf{\textit{To achieve sensor-agnostic robustness}}, we propose \textbf{RAWild}, a histogram-guided global-local tone mapping framework. Lightweight \textbf{\textit{differentiable B\'ezier curves}} first handle global tonal adjustment, producing physically plausible illumination correction without color distortion. The \textbf{\textit{Bilateral Grid}} then performs local, spatially-adaptive color adjustment in a compact 3D bilateral space~\cite{chen2016bilateral}, where learned affine transformations deliver efficient, edge-aware detail enhancement at negligible computational cost and preserve the high-frequency structures relevant for detection. Statistical priors extracted from the RAW pixels distribution jointly guide both the curve parameters and the grid coefficients, yielding a content-aware, scene-adaptive mapping that generalizes robustly across diverse lighting conditions. Extensive experiments demonstrate that our method achieves SOTA performance not only on multiple established RAW benchmark datasets but also under mixed-dataset training regimes spanning heterogeneous sensors and bit-depths, as well as more challenging robustness scenarios involving severe degradation and domain shift, as illustrated in \Cref{fig:radar}.

Our key contributions could be summarized as follows:
\begin{itemize}
\item We propose \textbf{RAWild}, a histogram-guided global-local tone mapping framework for
 sensor-agnostic RAW object detection. It decomposes sensor variations into a global B\'ezier curve for tonal correction and a spatially-adaptive Bilateral Grid for local color refinement.
\item We reparametrize both components with a delta-residual Bézier curve, and a gain–mixing decomposition of the Bilateral Grid that separates per-channel gain from color mixing, enabling stable training across heterogeneous RAW images.
\item Extensive experiments on six per-sensor and mixed-sensor real-capture RAW benchmarks and three datasets synthesized by our physics-based RAW simulation pipeline, spanning 10- to 24-bit sensors, demonstrate consistent SOTA performance.
\end{itemize}

\section{Related Work}
\paragraph{Camera RAW-based Object Detection.}
RAW data preserves linear photon measurements with high dynamic range and avoids the irreversible loss of ISP~\cite{karaimer2016software,delbracio2021mobile}, benefiting perception in low light~\cite{hong2021crafting,chen2018learning,jiang2025learning} and HDR scene reconstruction~\cite{mildenhall2022nerf,jin2024lighting,li2024chaos,kee2025removing}. Detection-oriented efforts either jointly learn task-aware ISPs with the detector~\cite{yu2021reconfigisp,wang2024adaptiveisp,li2024dualdn,kim2023paramisp}, adapt sRGB-pretrained models to RAW with adaptive modules~\cite{cui2024raw,huang2025dr}, or build RAW-native detectors spanning embedded~\cite{omid2017toward}, low-light~\cite{guo2025dark}, and diverse-condition settings \cite{xu2023toward,gamrian2025beyond,xie2026simrod,li2024efficient,li2025towards,li2025real}. These detectors remain optimized per sensor and are brittle to cross-sensor radiometric shifts.

\paragraph{Image Signal Processing with Bilateral Grid.}
Bilateral Grid~\cite{chen2007bilateral,chen2016bilateral} embeds pixels in a 3D spatial-intensity space for efficient edge-aware operations sliced back to full resolution; HDRnet~\cite{gharbi2017deep} first paired it with deep learning to predict spatially-varying color transforms, joining other learned ISP components~\cite{gharbi2016deep,heide2014flexisp}. Follow-ups cascade grids with neural 3D LUTs for retouching~\cite{kim2024image,conde2024nilut,zehtab2025efficient}, embed per-view grids in radiance fields to disentangle ISP effects~\cite{wang2024bilateral,wang2025unifying}, or combine them with learned tonal curves and color-name priors~\cite{serrano2024namedcurves,liu2024mlp,le2023gamutmlp}. We are the first to apply Bilateral Grids to RAW detection and couple them with a global B\'ezier curve to achieve physics-aware tonal mapping.

\paragraph{Sensor-Agnostic Computer Vision.}
Sensor-agnostic learning unifies heterogeneous inputs so one model generalizes across modalities, e.g., depth completion across LiDAR/ToF~\cite{park2024depth}, 3D reconstruction across fisheye and rolling-shutter cameras~\cite{wu20253dgut}, and video generation across diverse optics~\cite{zhang2025unified}. In the RAW/ISP domain, well-documented cross-sensor variation in spectral sensitivity~\cite{jiang2013camspec} and radiometric response~\cite{healey1994radiometric} has motivated camera-prior approaches such as unpaired RAW-to-RAW translation~\cite{perevozchikov2024rawformer}, parameter-aware forward/inverse ISPs~\cite{kim2023paramisp}, and metadata/EXIF-conditioned models~\cite{afifi2025time,zheng2023exif}. Unlike these, we target sensor-agnostic RAW \emph{detection}, addressing bit depth, spectral sensitivity, and exposure via a physics-guided global-local framework.

\section{Method}
\label{sec:method}

\begin{figure}[!tp]
    \centering
    \includegraphics[width=\textwidth]{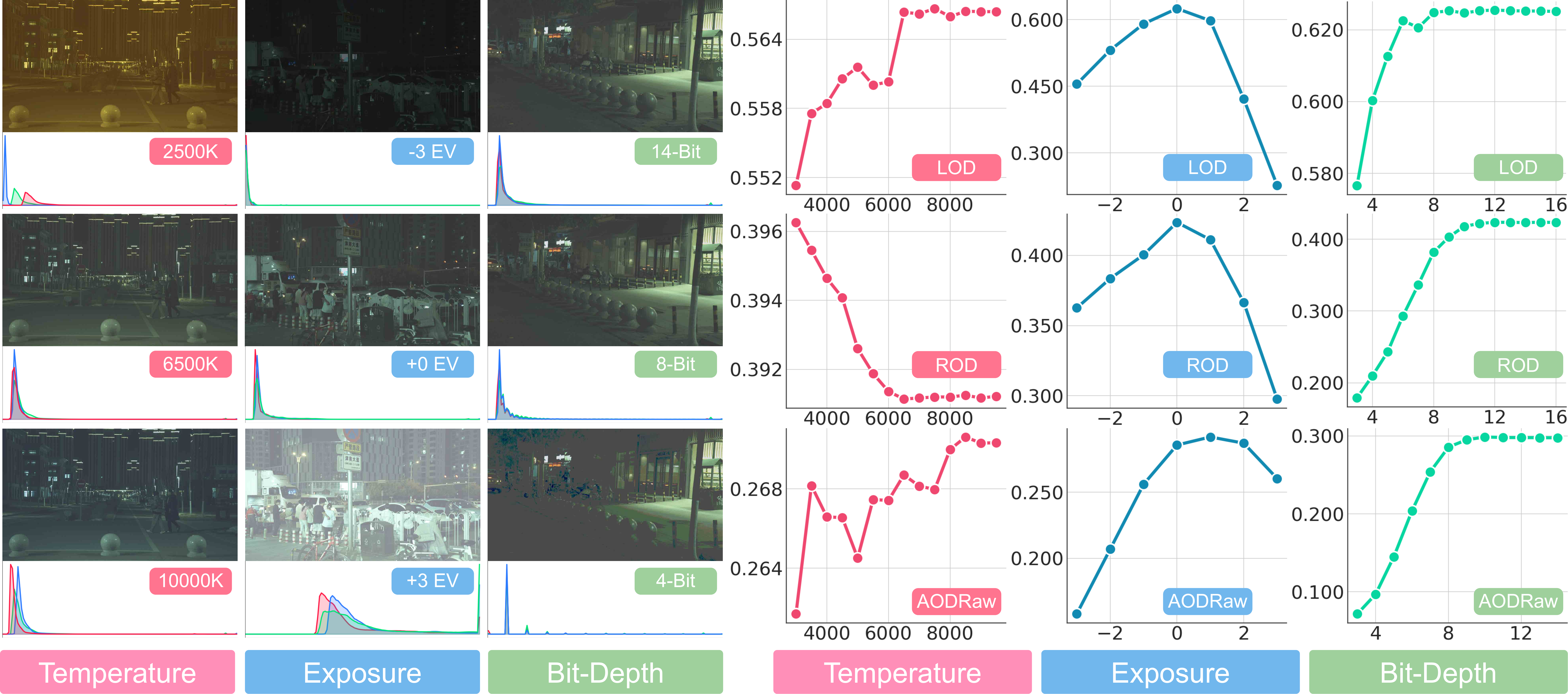}
    \caption{Sensitivity of RAW object detection to environmental and sensor-induced variations. Left: three AODRAW frames rendered under shifted illumination color temperature, exposure, and sensor bit depth, each paired with its pixel histogram. Right: mAP\@50 of a pretrained ResNet50 backbone with a linear RAW stem on LOD, ROD, and AODRAW as the three factors sweep across their ranges.}
    \label{fig:histogram}
\end{figure}

\subsection{Motivation}
\label{sec:motivation}

Many of the factors discussed in the introduction collapse along common axes. The illuminant spectrum and camera spectral response jointly control the per-channel color balance, i.e., color temperature and tint. Scene radiance, shutter time, and analog gain together scale the overall signal level, i.e., exposure. The sensor's non-linear response and ADC bit depth govern tonal compression and quantization at highlights and shadows. We train a linear-RAW-stem detector on three RAW benchmarks~\cite{hong2021crafting,xu2023toward,li2025towards} and evaluate it under controlled shifts along each axis. As shown in \Cref{fig:histogram}, detection accuracy drops sharply along all three once the input deviates from the training capture setting; a sensor-agnostic detector must therefore absorb variation along all of them. These distortions act at two scales: a global tonal shift from illuminant, exposure, and sensor non-linearity, and local color perturbations from mixed illumination, lens-shading falloff, and cross-channel crosstalk.

We factor the adapter into two physically meaningful operators: (i) a global per-channel curve $g$ that absorbs color temperature, exposure, and bit-depth shifts via a single monotone function, and (ii) a spatially- and luminance-adaptive color translation matrix $M$ that handles residual local errors. The output $y$ then takes the form:
\begin{equation}
y(p) \;=\; M\big(p,\, \ell(p)\big)\, g\big(I(p)\big),
\qquad \ell(p) \;=\; \frac{1}{3}\sum_{c} I_c(p),
\label{eq:adapter}
\end{equation}
where $I, y \in [0,1]^{3\times H\times W}$ are the normalized linear RAW input (after black-level subtraction) and the backbone-ready output. Crucially, $\ell(p)$ is computed from the unmapped $I$ rather than $g(I)$, so its luminance axis retains a fixed physical meaning across sensors and training stages. We parameterize $g$ as a degree-$n$ B\'ezier curve (\Cref{sec:bezier}) and store $M$ as a low-resolution Bilateral Grid (\Cref{sec:grid}); both are jointly predicted by a shared histogram-guided network (\Cref{sec:arch}).

\begin{figure}[!tp]
    \centering
    \includegraphics[width=\textwidth]{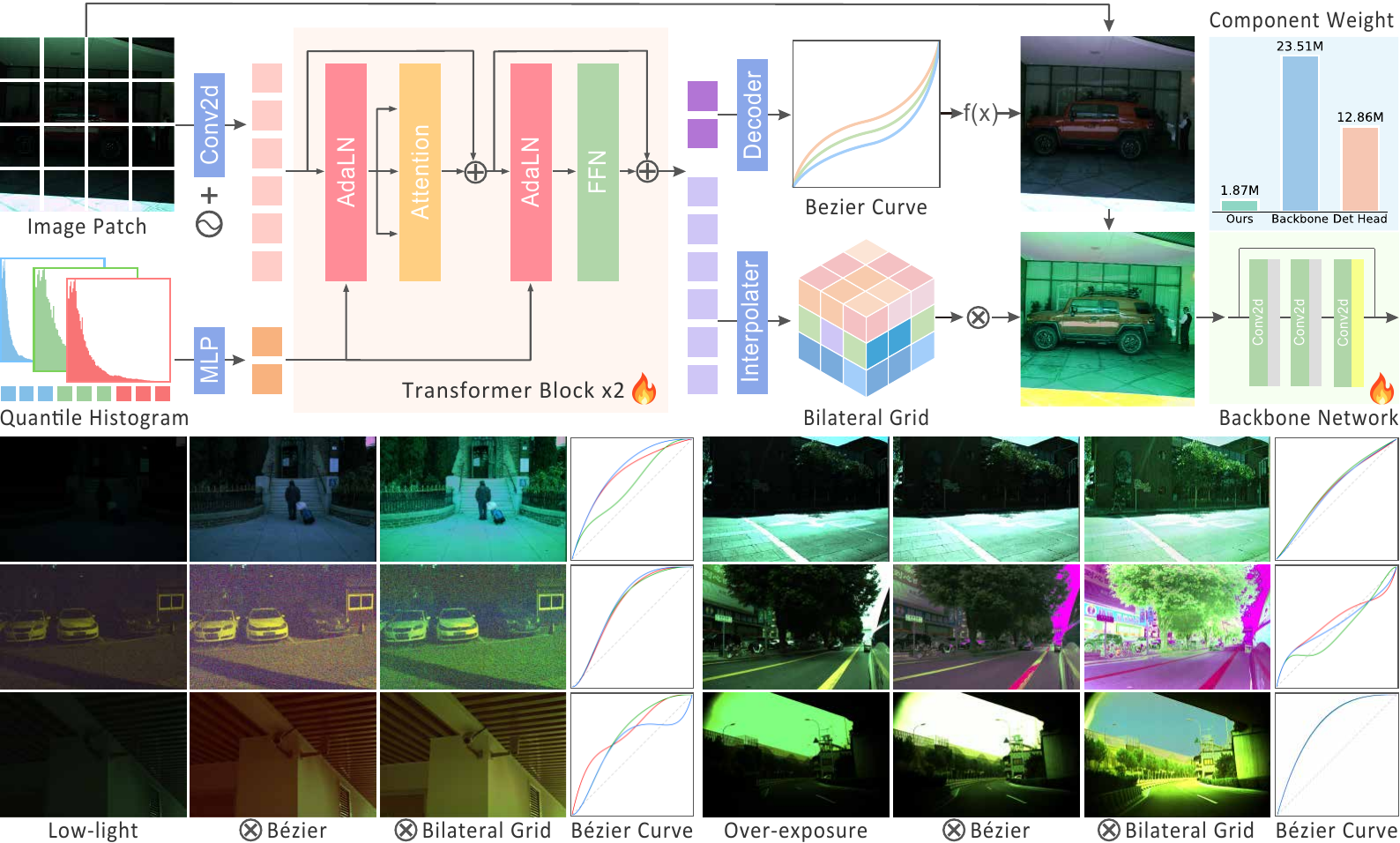}
    \caption{Overall architecture of the proposed method. A RAW image and its quantile histogram drive transformer blocks via AdaLN, jointly predicting a global B\'ezier curve $f(x)$ and a Bilateral Grid for local color correction. Qualitative examples on low-light and over-exposed inputs show the B\'ezier-only output, the full Bilateral Grid output, and the learned curve.}
    \label{fig:pipeline}
\end{figure}

\subsection{Global Differentiable B\'ezier Curve}
\label{sec:bezier}

As illustrated on top of \Cref{fig:pipeline}, the global tone curve $g$ maps the linear RAW signal through a smooth, per-channel monotone function, absorbing exposure, sensor non-linear response, and per-channel sensitivity into $3(n{-}1)$ scalar coefficients. We parameterize $g$ as a degree-$n$ B\'ezier curve per channel \cite{serrano2024namedcurves}, which is pointwise bounded, monotone under ordered control points, and pinned to pure black and white at its endpoints. Let $\{p_{c,i}\}_{i=0}^{n}\subset[0,1]$ denote the $n{+}1$ control points for channel $c\in\{R,G,B\}$. Given $t=I_c(p)\in[0,1]$, the curve evaluates pointwise as:
\begin{equation}
g_c(t) \;=\; \sum_{i=0}^{n} B_{n,i}(t)\, p_{c,i},
\qquad
B_{n,i}(t) \;=\; \binom{n}{i}(1-t)^{n-i}\, t^{i},
\label{eq:bezier}
\end{equation}
where $\{B_{n,i}\}_{i=0}^{n}$ are the Bernstein basis polynomials. Since $B_{n,i}(t)\geq 0$ and $\sum_{i}B_{n,i}(t)=1$, each $g_c(t)$ is a convex combination of its control points and stays within $[\min_i p_{c,i},\max_i p_{c,i}]$. The derivative of $g_c$ admits the closed form:
\begin{equation}
g_c'(t) \;=\; n\sum_{i=0}^{n-1} B_{n-1,i}(t)\,\bigl(p_{c,i+1}-p_{c,i}\bigr),
\label{eq:bezier_deriv}
\end{equation}
which is a convex combination of adjacent control-point differences. It stays non-negative under ordered control points $p_{c,0}\le\cdots\le p_{c,n}$ and renders $g_c$ monotonically non-decreasing. To regularize the predicted curve toward this ordered regime, we anchor the interior points on the uniform ramp $\bar p_{c,i}=i/n$ and apply a bounded residual that keeps each update a bounded perturbation of identity as:
\begin{equation}
p_{c,i} \;=\; \mathrm{clip}_{[0,1]}\Bigl(\tfrac{i}{n} \;+\; \tfrac{1}{2}\tanh(\delta_{c,i})\Bigr),
\qquad i=1,\dots,n-1.
\label{eq:bezier_ctrl}
\end{equation}
Our B\'ezier parameterization is both more expressive than a per-channel gamma~\cite{xie2026simrod}, whose single scalar cannot fit the S-shaped response of modern sensors, and more constrained than free-form polynomial bases~\cite{guo2025dark}, which provide no endpoint or monotonicity guarantees and may crush or invert tones near black and white. Because $\delta_{c,i}$ is predicted per image from sensor-dependent statistics in \Cref{sec:arch}, $g$ specializes adaptively across sensors. A wide-dynamic-range Bayer sensor receives an aggressively compressive curve, whereas a narrow-range low-bit sensor receives a near-linear one. The $3(n{-}1)$ learned degrees of freedom thus absorb exposure, gamma, and per-channel sensitivity differences without any sensor-specific architecture.

\subsection{Local Bilateral Grid Color Transform}
\label{sec:grid}

The global curve leaves spatially-varying color shifts intact. A single RAW frame routinely contains mixed illumination, lens-shading falloff, and sensor-dependent crosstalk. Detection does not require a colorimetrically faithful image, but benefits from enhanced local contrast and balanced color that expose object boundaries and textures to the backbone. We therefore apply a spatially-adaptive $3{\times}3$ color transform, represented as a bilateral grid following~\cite{chen2007bilateral,gharbi2017deep}, as depicted in the bottom of \Cref{fig:pipeline}.

\paragraph{Grid Representation and Trilinear Slicing.}
The grid $\mathbf{G}\in\mathbb{R}^{9\times G_d\times G_h\times G_w}$ stores the nine coefficients of a $3{\times}3$ matrix at each cell, with spatial resolution $(H/16){\times}(W/16)$ and $G_d{=}8$ luminance bins. Each full-resolution pixel $p$ recovers its coefficients by trilinear interpolation into $\mathbf{G}$ at the continuous coordinate $(x G_w/W, y G_h/H, \ell(p) G_d)$,
\begin{equation}
m(p) \;=\; \!\!\sum_{(d,h,w)\in\mathcal{N}(p)}\!\! w_d(p)\,w_h(p)\,w_w(p)\; \mathbf{G}_{:,d,h,w} \;\;\in\;\mathbb{R}^{9},
\label{eq:slice}
\end{equation}
where $\mathcal{N}(p)$ is the eight enclosing grid vertices and $(w_d,w_h,w_w)$ are the linear interpolation weights. The sampled vector reshapes into a $3{\times}3$ matrix $M(p)$ and applies pointwise as:
\begin{equation}
y(p) \;=\; \mathrm{clip}_{[0,1]}\bigl(M(p)\,g\bigl(I(p)\bigr)\bigr).
\label{eq:render}
\end{equation}
Prediction operates at grid resolution and rendering is $\mathcal{O}(HW)$, so the luminance axis lets pixels at the same location but different brightness sample different matrices without per-pixel prediction cost.

\paragraph{Structured Matrix Parameterization.}
A free $3{\times}3$ prediction at every cell is fragile in practice. Sensor-to-sensor color differences are dominated by per-channel gain absorbing white balance, RAW-domain exposure, and spectral-sensitivity scaling, which can vary by a factor of three or more across illuminants and sensors. Channel mixing absorbs residual cross-spectral overlap and color-filter leakage, and is a small correction of at most $10$ to $20\%$ relative to the dominant channel~\cite{karaimer2016software}. An unconstrained $3{\times}3$ conflates these two effects at markedly different magnitudes. Gradients on the dominant gain contaminate the off-diagonals, and early-iteration matrices that are ill-conditioned or sign-flipped destroy the pretrained backbone's features. We therefore parameterize every cell of $\mathbf{G}$ as:
\begin{equation}
M \;=\; D\,(I + A),
\qquad
D = \operatorname{diag}(d),\;
d = \exp\bigl(\tanh(\hat d)\bigr),\;
A_{ij\mid i\neq j} = k\tanh(\hat a_{ij}),
\label{eq:dia}
\end{equation}
where $\hat d\in\mathbb{R}^3$ and $\hat a\in\mathbb{R}^6$ are the nine raw coefficients emitted by the prediction network, with fixed activation bound $k{=}0.05$. The matrix takes the explicit form:
\begin{equation}
M \;=\;
\begin{bmatrix}
d_1 & d_1 A_{12} & d_1 A_{13} \\
d_2 A_{21} & d_2 & d_2 A_{23} \\
d_3 A_{31} & d_3 A_{32} & d_3
\end{bmatrix},
\label{eq:dia_expanded}
\end{equation}
in which each row is a gain $d_i$ multiplied by a unit-diagonal mixing vector. The diagonal entries lie in $[e^{-1},e^{1}]\approx[0.37,2.72]$, so $d$ neither collapses to zero nor flips sign, and color channels retain their physical orientation throughout training. With $|A_{ij}|\le k$, the matrix $I+A$ is strictly diagonally dominant and invertible, and so is $M=D(I+A)$, which prevents the detector backbone from observing degenerate inputs. Left-multiplication by $D$ further ensures that $A$ encodes mixing ratios relative to each row's gain rather than absolute channel deltas, so globally rescaling a channel adjusts only $d_i$ and leaves $A$ invariant. This decoupling eliminates the gradient interference between exposure and color-mixing corrections that plagues unconstrained $3{\times}3$ predictions.

Zero-initializing the grid head gives $\hat d{=}\hat a{=}\mathbf{0}$, hence $d{=}\mathbf{1}$, $A{=}\mathbf{0}$, and $M{=}I$. Combined with the identity-initialized B\'ezier curve of~\Cref{eq:bezier_ctrl}, the full adapter~\Cref{eq:adapter} is exactly the identity at initialization and thereafter learns a bounded, well-conditioned residual, allowing a single network to train jointly across heterogeneous sensors.

\begin{figure}[!tp]
    \centering
    \includegraphics[width=\textwidth]{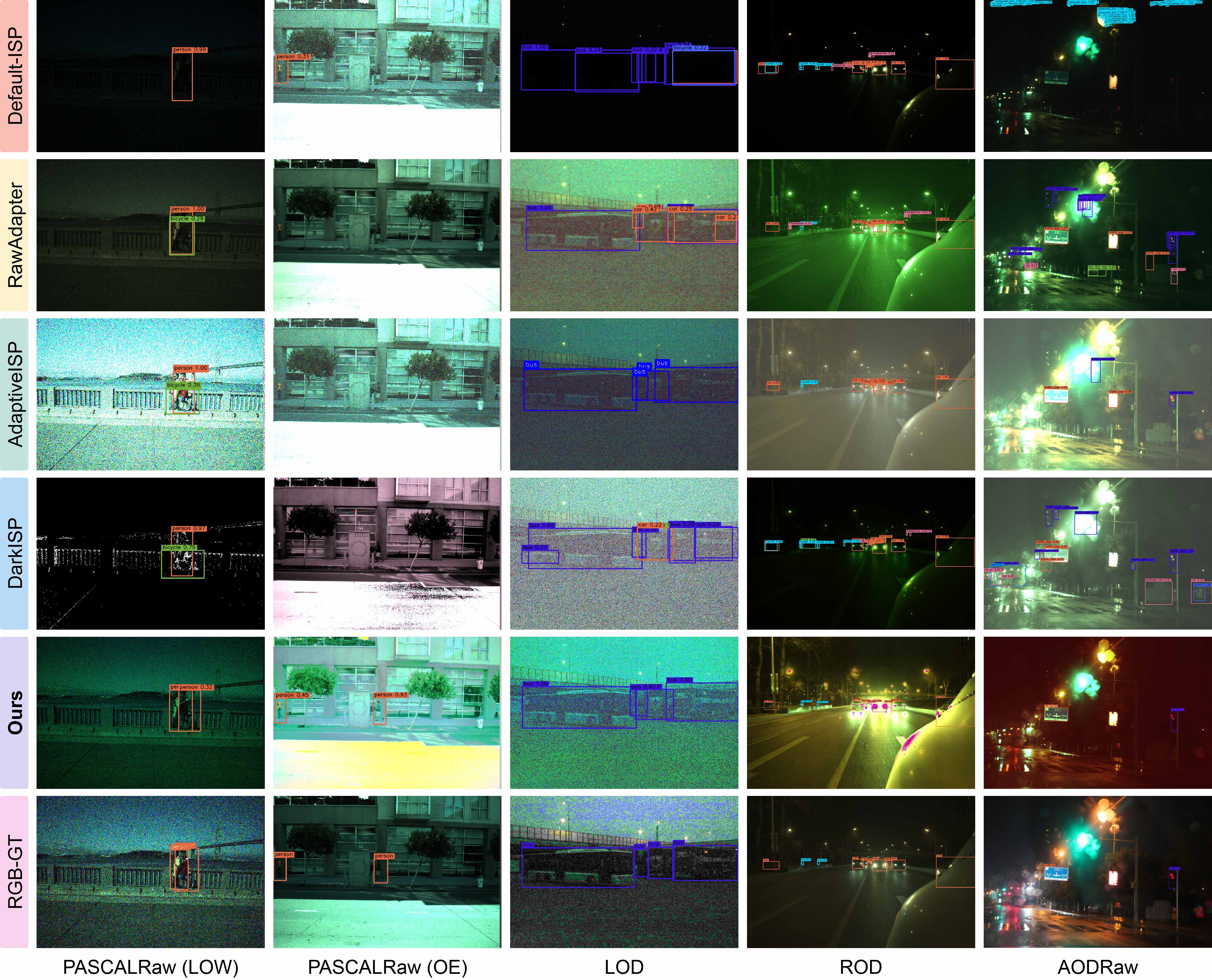}
    \caption{Qualitative comparison of object detection on PASCALRAW~\cite{omid2017toward}, LOD~\cite{hong2021crafting}, ROD~\cite{xu2023toward}, and AODRAW~\cite{li2025towards}. Our method produces more robust and accurate detections than the baselines. Zoom in for details.}
    \label{fig:det_result}
\end{figure}

\subsection{Histogram-Guided Curve and Grid Prediction}
\label{sec:arch}

The B\'ezier residual $\delta$ and the Bilateral Grid $\mathbf{G}$ are predicted by a single shared network from the normalized linear RAW input $I$. Four stride-$2$ $3{\times}3$ convolutions with channel progression $3{\to}32{\to}64{\to}128{\to}D$ produce a token map of shape $D\times G_h\times G_w$, which is augmented with fixed 2D sin--cos positional encodings~\cite{vaswani2017attention} and refined by two transformer blocks.

To inject sensor-level priors, we compute a differentiable per-channel quantile descriptor $\mathbf{q}\in\mathbb{R}^{3\times Q}$ with $Q{=}64$ entries per channel as a differentiable surrogate of the channel histogram, where for each rank $r\in\{1,\ldots,Q\}$ the entry $q_r$ is the $r/Q$-th empirical quantile of the pixel intensities, obtained via soft sorting for end-to-end differentiability, and project it through a small MLP to produce per-layer AdaLN~\cite{peebles2023scalable} scale and shift parameters. These modulate the attention and FFN activations of each block, so the same network adapts to sensor statistics such as bit depth, white-balance bias, exposure regime, and dynamic-range occupancy. The descriptor is equivariant to monotone rescalings of the signal, so relative spacings between successive $q_r$ rather than absolute values encode distribution shape, yielding robustness across heterogeneous bit depths and gains.

A $1{\times}1$ convolution expands each spatial token into $G_d\cdot 9$ raw grid coefficients, reshaped and passed through~\Cref{eq:dia} to yield $\mathbf{G}$. A linear head maps the average-pooled token features to the B\'ezier residual $\delta$, combined with the uniform anchor via~\Cref{eq:bezier_ctrl}. Only the convolutional stem depends on $HW$, so the adapter adds a near-constant per-image cost across detector architectures.

\section{Experiment}
\label{sec:experiment}

\paragraph{Implementation Details}
We implement our models in PyTorch with MMDetection. We loaded pretrained RetinaNet for all baseline methods, finetuned with SGD (lr $1\times10^{-3}$, momentum 0.9, weight decay $5\times10^{-4}$) for 20 epochs at batch size 4, with a 500 step linear warmup. Augmentation consists of resizing and random horizontal flipping (p$=$0.5). ~\Cref{fig:pipeline} (upper right) reports the parameter breakdown of the network components, and \Cref{tab:inference_cost} compares inference time and memory against representative baselines using a NVIDIA H100 GPU.

\begin{table}[!tp]
\centering
\caption{Evaluation of object detection performance in mAP@50 and @75 on PASCALRAW Low/Normal/Overexposed, LOD, ROD, and AODRAW. Models use a ResNet50~\cite{he2016deep} or a SwinTransformer~\cite{liu2021swin} backbone.}
\label{tab:obect_detection}
\setlength{\tabcolsep}{3pt}
\resizebox{\textwidth}{!}{
\begin{tabular}{l|l cc cc cc cc cc cc}
\toprule
\multicolumn{2}{l}{} & \multicolumn{2}{c}{PAS.LOW} & \multicolumn{2}{c}{PAS.NM} & \multicolumn{2}{c}{PAS.OE}
& \multicolumn{2}{c}{LOD} & \multicolumn{2}{c}{ROD} & \multicolumn{2}{c}{AODRAW} \\
\cmidrule(lr){3-4} \cmidrule(lr){5-6} \cmidrule(lr){7-8}
\cmidrule(lr){9-10} \cmidrule(lr){11-12} \cmidrule(lr){13-14}
& Method & @50 & @75 & @50 & @75 & @50 & @75 & @50 & @75 & @50 & @75 & @50 & @75 \\
\midrule
\rowcolor{default}
\multirow{8}{*}{\cellcolor{white}\rotatebox[origin=c]{90}{ResNet-50}}
& Linear-RAW & 0.7668 & 0.5793 & 0.8661 & 0.7273 & 0.8747 & 0.7298 & 0.5710 & 0.3713 & 0.3041 & 0.1963 & 0.2301 & 0.1485 \\
& Default-ISP & 0.8390 & 0.6676 & 0.8959 & 0.7328 & 0.8967 & 0.7333 & 0.5880 & 0.3549 & 0.4178 & 0.2709 & 0.3241 & \thd 0.2063 \\
& ReconfigISP & 0.8562 & 0.6662 & 0.8949 & 0.7286 & \thd 0.8989 & 0.7263 & 0.5383 & 0.2903 & 0.3436 & 0.2137 & 0.2896 & 0.1763 \\
& RAW-Adapter   & 0.8647 & 0.6935 & 0.8942 & 0.7270 & 0.8987 & 0.7333 & 0.6082 & 0.3285 & 0.3950 & 0.2560 & 0.2934 & 0.1878 \\
& AdaptiveISP & \sed 0.8871 & \thd 0.7293 & \sed 0.9014 & \fst 0.7872 & \sed 0.9011 & \sed 0.7636 & \thd 0.6142 & \thd 0.4071 & \thd 0.4379 & \thd 0.2848 & \thd 0.3382 & \sed 0.2131 \\
& Dark-ISP & 0.7139 & 0.4663 & 0.8940 & 0.7550 & 0.8840 & 0.6890 & 0.5560 & 0.3630 & 0.4090 & 0.2710 & 0.3240 & 0.2060 \\
& Dr.RAW & \thd 0.8865 & \sed 0.7305 & \thd 0.8983 & \thd 0.7639 & 0.8982 & \thd 0.7624 & \sed 0.6408 & \fst 0.4399 & \sed 0.4720 & \sed 0.3160 & \fst 0.3630 & \sed 0.2310 \\
& \textbf{Ours} &  \fst \textbf{0.8910} &  \fst \textbf{0.7330} &  \fst \textbf{0.9020} &  \sed \textbf{0.7690} &  \fst \textbf{0.9020} & \fst \textbf{0.7690} &  \fst \textbf{0.6909} &  \sed \textbf{0.4363} &  \fst \textbf{0.5541} &  \fst \textbf{0.3752} &  \sed \textbf{0.3623} &  \fst \textbf{0.2311} \\
\midrule
\rowcolor{default}
\multirow{8}{*}{\cellcolor{white}\rotatebox[origin=c]{90}{Swin-Transformer}}
& Linear-RAW & 0.5517 & 0.3336 & 0.8707 & 0.6882 & 0.8733 & 0.6196 & 0.5279 & 0.2828 & 0.3251 & 0.2120 & 0.1241 & 0.0768 \\
& DefaultISP & 0.8581 & \thd 0.7156 & \sed 0.9023 & \thd 0.7769 & 0.8966 & 0.7410 & \sed 0.6978 & \thd 0.5015 & 0.4885 & 0.3119 & \thd 0.3428 & \thd 0.2308 \\
& ReconfigISP & \thd 0.8803 & 0.6999 & 0.8946 & 0.7165 & \sed 0.9017 & \sed 0.7618 & 0.5935 & 0.3746 & 0.4051 & 0.2527 & 0.2977 & 0.1966 \\
& RAW-Adapter  & 0.8727 & 0.7121 & \thd 0.9017 & \fst 0.7869 & 0.8967 & 0.7546 & 0.6289 & 0.4149 & 0.4386 & 0.2869 & 0.3035 & 0.1985 \\
& AdaptiveISP & 0.7830 & 0.5516 & 0.8704 & 0.7239 & 0.8883 & 0.7308 & 0.5990 & 0.4581 & 0.4440 & 0.2711 & \sed 0.3661 & \sed 0.2471 \\
& Dark-ISP & 0.7233 & 0.4852 & 0.9010 & 0.7672 & \thd 0.8980 & \thd 0.7556 & 0.6303 & 0.4146 & \sed 0.5039 & \sed 0.3273 & 0.3398 & 0.2233 \\
& Dr.RAW & \sed 0.8814 & \sed 0.7183 & 0.8986 & 0.7591 & 0.8958 & 0.7190 & \thd 0.6950 & \fst 0.5266 & \thd 0.4933 & \thd 0.3192 & 0.3268 & 0.2160 \\
& \textbf{Ours} & \fst \textbf{0.9083} & \fst \textbf{0.7406} & \fst \textbf{0.9342} & \sed \textbf{0.7792} & \fst \textbf{0.9313} & \fst \textbf{0.7921} & \fst \textbf{0.6986} & \sed \textbf{0.5057} & \fst \textbf{0.5191} & \fst \textbf{0.3467} & \fst \textbf{0.4394 }& \fst \textbf{0.3263} \\

\bottomrule
\end{tabular}%
}
\end{table}

\begin{table*}[t]
\centering
\footnotesize
\begin{minipage}[t]{0.75\textwidth}
\vspace{0pt}
\centering
\caption{Evaluation on synthetic and real mixed-sensor datasets.}
\label{tab:hybrid_dataset}
\setlength{\tabcolsep}{1.3pt}
\begin{tabular}{lcccccc|cccc}
\toprule
& \multicolumn{2}{c}{PAS (Syn)}
& \multicolumn{2}{c}{LOD (Syn)}
& \multicolumn{2}{c|}{ROD (Syn)}
& \multicolumn{2}{c}{PAS \& LOD}
& \multicolumn{2}{c}{Multi-RAW} \\
\cmidrule(lr){2-3}
\cmidrule(lr){4-5}
\cmidrule(lr){6-7}
\cmidrule(lr){8-9}
\cmidrule(lr){10-11}
Method & @50 & @75 & @50 & @75 & @50 & @75 & @50 & @75 & @50 & @75 \\
\midrule
\rowcolor{default}
Linear-RAW & 0.8921 & 0.7295 & 0.5310 & 0.3297 & 0.3497 & 0.2273 & 0.6057 & 0.4175 & 0.2539 & 0.1227 \\
RAW-Adapter  & 0.8869 & 0.7233 & \thd 0.5874 & \thd 0.3945 & 0.3637 & 0.2397 & \thd 0.6540 & \thd 0.4370 & 0.2334 & 0.1025 \\
Dr.RAW & \sed 0.8949 & \thd 0.7324 & \sed 0.5924 & \sed 0.4164 & \sed 0.4234 & \sed 0.2821 & \sed 0.6660 & \sed 0.4750 & \thd 0.2572 & \sed 0.1195 \\
Dark-ISP & \thd 0.8888 & \sed 0.7334 & 0.5837 & 0.3390 & \thd 0.3703 & \thd 0.2403 & 0.6420 & 0.4360 & \sed 0.2603 & \thd 0.1063 \\
\textbf{Ours} & \fst \textbf{0.8953} & \fst \textbf{0.7343} & \fst \textbf{0.6646} & \fst \textbf{0.4602} & \fst \textbf{0.4291} & \fst \textbf{0.3223} & \fst \textbf{0.6865} & \fst \textbf{0.4896} & \fst \textbf{0.3528} & \fst \textbf{0.1336} \\
\bottomrule
\end{tabular}
\end{minipage}
\hspace{-2mm}
\begin{minipage}[t]{0.25\textwidth}
\vspace{0pt}
\centering
\caption{Efficiency.}
\label{tab:inference_cost}
\setlength{\tabcolsep}{1.3pt}
\begin{tabular}{lcc}
\toprule
Inference & \multicolumn{2}{c}{Time\&Mem.} \\
\cmidrule(r){1-1} \cmidrule(lr){2-3}
Method & ms & GB \\
\midrule
\rowcolor{default}
Linear-RAW & 7.20 & 0.23 \\
RAW-Adapter  & \fst 10.00 & \fst 0.24 \\
Dr.RAW & \sed 10.52 & \sed 0.53 \\
Dark-ISP & 20.45 & \thd 0.99 \\
\textbf{Ours} & \thd \textbf{10.70} & \fst \textbf{0.24} \\
\bottomrule
\end{tabular}
\end{minipage}
\end{table*}

\paragraph{Baseline Methods.}
We compare our method with default ISP and recent task-oriented ISP approaches, including ReconfigISP~\cite{yu2021reconfigisp}, RAW-Adapter ~\cite{cui2024raw}, AdaptiveISP~\cite{wang2024adaptiveisp}, Dark-ISP~\cite{guo2025dark}, and Dr.RAW~\cite{huang2025dr}.

\paragraph{Benchmarks.}
For object detection on RAW images, we use the 12-bit PASCALRAW dataset~\cite{omid2017toward} and its lowlight (\emph{LOW}), normal (\emph{NM}), and overexposure (\emph{OE}) variants~\cite{cui2024raw}. We further adopt the 14-bit AODRAW~\cite{li2025towards}, 16-bit LOD~\cite{hong2021crafting}, and 24-bit ROD~\cite{xu2023toward} datasets, along with the Multi-RAW dataset~\cite{li2024efficient} that contains 10-, 12-, 16-, and 24-bit RAW images captured by diverse sensors. We additionally construct a mixed-sensor dataset from PASCALRAW and LOD, and synthesize multi-sensor dataset, with synthesis details provided in the Appendix~\ref{sec:raw_aug}. Additional experiments of RAW image segmentation are provided in Appendix~\ref{app:additional_segmentation}.

\paragraph{Metric.}
We report object detection accuracy in mAP@50 and mAP@75. Best results in experiments are shaded as \colorbox{fst}{first}, \colorbox{sed}{second}, and \colorbox{thd}{third}. \colorbox{default}{Linear-RAW} denotes direct integration of RAW images.

\subsection{Evaluation on Per-Sensor Benchmarks}
\Cref{tab:obect_detection} reports mAP@50 and mAP@75 across the six per-sensor benchmarks under both a ResNet-50 and a Swin-Transformer backbone. Our adapter achieves superior or competitive performance on all settings. The improvement is more pronounced beyond the conventional 12-bit setting; on ROD and LOD, ResNet-50 mAP@50 improves by 17.4\% and 7.8\% over the strongest baseline approaches. Across the three PASCALRAW exposure variants constructed by~\cite{cui2024raw}, the model recovers nearly identical mAP@50 on LOW, NM, and OE, suggesting that the bounded global curve absorbs exposure variation rather than overfitting to a particular setting. \Cref{fig:det_result} shows fewer missed and mislocalized boxes under saturated highlights and severely underexposed regions.

\subsection{Evaluation on Mixed-Sensor Benchmarks}
\Cref{tab:hybrid_dataset} examines how each method behaves when training and evaluation span heterogeneous sensor captures. Our method attains the highest mAP@50 among task-oriented ISPs in both synthetic and real-world settings, with relative gains of 6.1\% on LOD (Syn), 3.1\% on PASCALRAW \& LOD, and 9.8\% on MultiRAW over the strongest baselines. The margin is largest on MultiRAW, which spans the widest range of bit depths and capture devices in our evaluation. On such heterogeneous mixes, previous pipelines tend to fit their parameters to a dominant sensor mode, whereas our histogram-conditioned prediction adapts to per-image statistics through the global B\'ezier curve and bilateral grid, which generalizes more reliably across sensor configurations.

\begin{figure}[!tp]
    \centering
    \begin{subfigure}[t]{0.496\linewidth}
        \centering
        \includegraphics[
            width=\linewidth
        ]{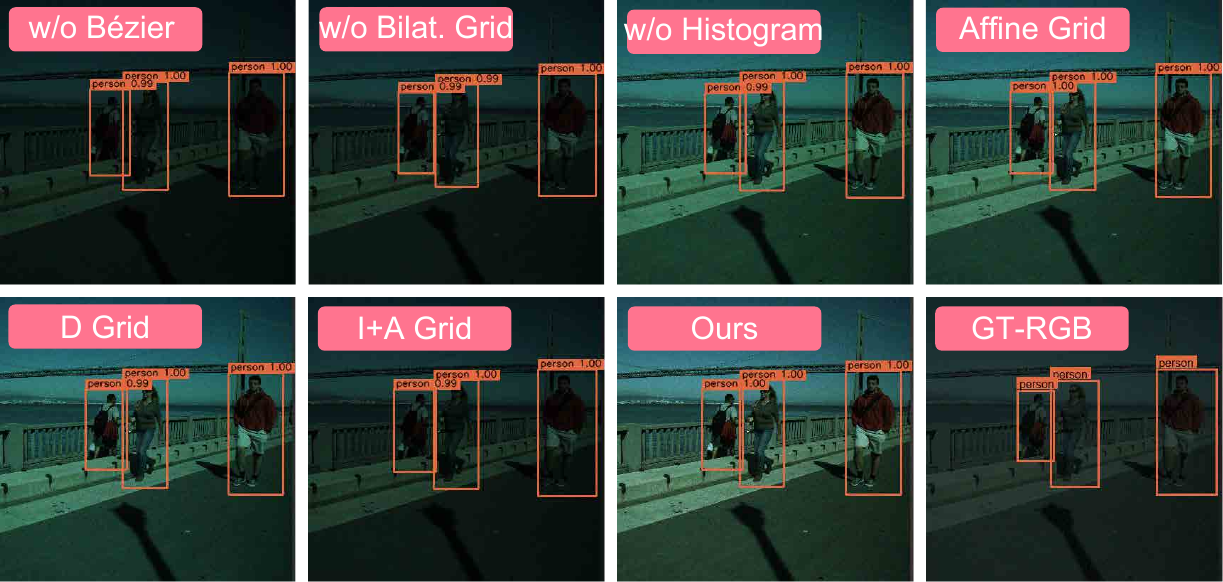}
        \vspace{-1.5em}
        \caption{Low-light}
        \label{fig:ablation_pascal_low}
    \end{subfigure}
    \hfill
    \begin{subfigure}[t]{0.496\linewidth}
        \centering
        \includegraphics[
            width=\linewidth
        ]{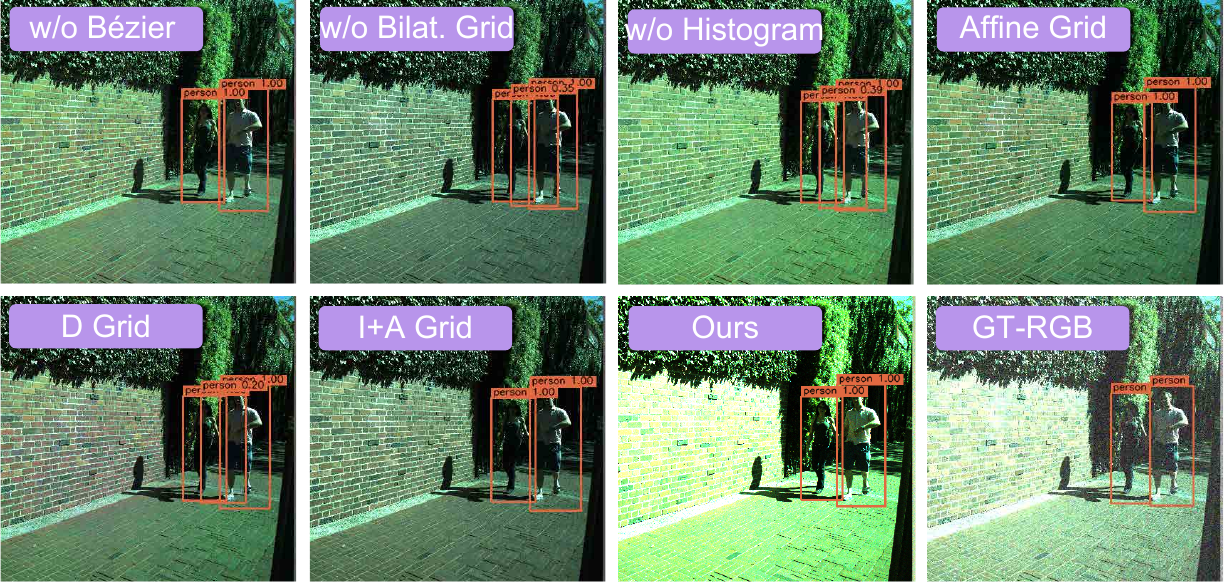}
        \vspace{-1.5em}
        \caption{Over-exposure}
        \label{fig:ablation_pascal_oe}
    \end{subfigure}
    \vspace{-0.5em}
    \caption{
        Visualization of ablation studies on PASCALRAW lowlight and overexposure variants. Each case presents transformed images of design choices shown in \Cref{tab:ablation_histogram}, \Cref{tab:ablation_bezier}, and \Cref{tab:ablation_grid}.
    }
    \label{fig:ablation_pascal}
\end{figure}

\begin{table*}[!tp]
\centering
\setlength{\tabcolsep}{3pt}
\renewcommand{\arraystretch}{1.0}
\begin{minipage}[t]{0.3\linewidth}
\vspace{0pt}
\caption{Ablation of histogram conditioning mechanisms.}
\label{tab:ablation_histogram}
\centering
{\footnotesize
\begin{tabularx}{\linewidth}{Xcc}
\toprule
Ablation & @50 & @75 \\
\midrule
No Histogram & 0.6838 & 0.4241 \\
Prefix Condition & 0.6397 & 0.4123 \\
\rowcolor{fst}
AdaLN Condition & \textbf{0.6909} & \textbf{0.4363} \\
\bottomrule
\end{tabularx}}
\label{tab:cond}
\end{minipage}%
\hspace{0.015\linewidth}%
\begin{minipage}[t]{0.28\linewidth}
\vspace{0pt}
\caption{Ablation study of global Bézier curve design.}
\label{tab:ablation_bezier}
\centering
{\footnotesize
\begin{tabularx}{\linewidth}{Xcc}
\toprule
Ablation & @50 & @75 \\
\midrule
No Bézier & 0.6650 & 0.4133 \\
Vanilla Bézier & 0.6511 & 0.4343 \\
\rowcolor{fst}
Delta Bézier & \textbf{0.6909} & \textbf{0.4363} \\
\bottomrule
\end{tabularx}}
\label{tab:bez}
\end{minipage}%
\hspace{0.015\linewidth}%
\begin{minipage}[t]{0.38\linewidth}
\vspace{0pt}
\caption{Ablation of bilateral grid.}
\label{tab:ablation_grid}
\centering
\renewcommand{\arraystretch}{0.85}
{\footnotesize
\begin{tabularx}{\linewidth}{Xcc}
\toprule
Ablation & @50 & @75 \\
\midrule
No Bilateral Grid & 0.6691 & 0.4346 \\
Vanilla Bilateral Grid & 0.6673 & 0.4149 \\
$\mathbf{D}$ & 0.6589 & 0.4461 \\
$\mathbf{I}{+}\mathbf{A}$ & 0.6646 & 0.4118 \\
\rowcolor{fst}
$\mathbf{D}(\mathbf{I}{+}\mathbf{A})$ & \textbf{0.6909} & \textbf{0.4363} \\
\bottomrule
\end{tabularx}}
\label{tab:grid}
\end{minipage}
\vspace{-1em}
\end{table*}

\section{Ablation Study}
\label{sec:ablation}
Ablations are conducted on LOD dataset~\cite{hong2021crafting} with the same ResNet-50 backbone. \Cref{fig:ablation_pascal} visualizes the transformed linear images from each ablated module or design.

\begin{wrapfigure}{r}{0.5\textwidth}
    \centering
    \begin{subfigure}[t]{0.48\linewidth}
        \centering
        \includegraphics[width=\linewidth]{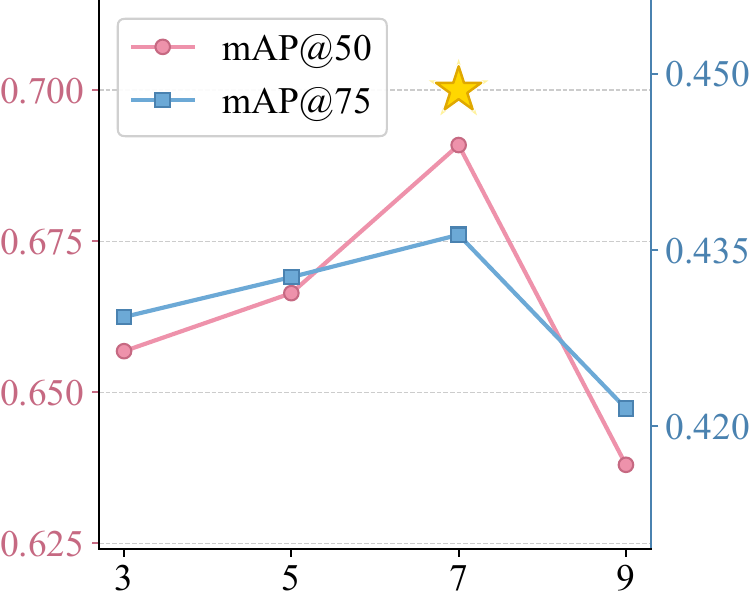}
        \caption{Points in B\'ezier curve}
        \label{fig:ablation_control_points}
    \end{subfigure}
    \hfill
    \begin{subfigure}[t]{0.48\linewidth}
        \centering
        \includegraphics[width=\linewidth]{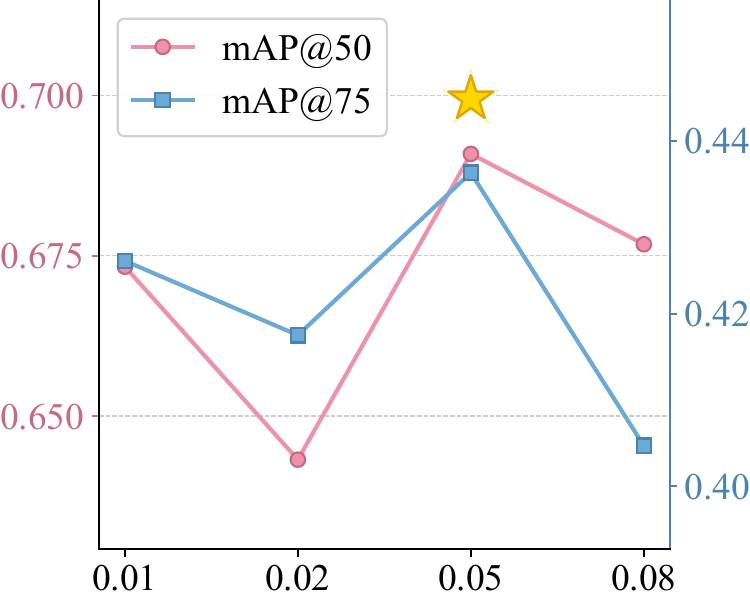}
        \caption{$k$ in Bilateral Grid}
        \label{fig:ablation_dia_k}
    \end{subfigure}
    \vspace{-0.5em}
    \caption{Ablation on grid activation $k$ and the number of B\'ezier control points. Star is our choice.}
    \vspace{-2em}
    \label{fig:ablation_dia}
\end{wrapfigure}

\paragraph{Effect of Histogram Conditioning}
\Cref{tab:cond} compares ways of injecting the input histogram. AdaLN modulation outperforms both removing the histogram and routing it through a prefix token, consistent with the histogram acting as a sensor-level prior that is most useful when it reshapes every block rather than competing with spatial tokens for attention.

\paragraph{Effect of Global B\'ezier Curve}
\Cref{tab:bez} compares forms of the global tone curve. Our delta parameterization outperforms both removing the curve and a vanilla B\'ezier that emits absolute control points, with the vanilla variant falling slightly below the no-curve baseline. This is consistent with the head benefiting from a uniform anchor rather than having to learn the identity transform from scratch on every input. \Cref{fig:ablation_control_points} further shows that accuracy plateaus once the curve has enough degrees of freedom to express the typical sensor response.

\paragraph{Effect of Spatial Bilateral Grid}
\Cref{tab:grid} disentangles the diagonal-mixing decomposition $\mathbf{D}(\mathbf{I}{+}\mathbf{A})$. The full form is best, whereas predicting only $\mathbf{D}$ or only $\mathbf{I}{+}\mathbf{A}$ falls even below the no-grid baseline, suggesting the gain comes from decoupling per-channel exposure and cross-channel mixing rather than from an increase in parameter count. \Cref{fig:ablation_dia_k} shows that the grid activation bound $k$ defined in \Cref{eq:dia} is stable across moderate values and degrades at the extremes where the grid approaches the identity or becomes nearly singular.

\section{Conclusion}
We address sensor-agnostic RAW object detection, where differences in bit depth, spectral response, and exposure across cameras prevent a single detector from generalizing. We factor the adapter into a global B\'ezier tone curve and a local bilateral grid of structured $\mathbf{D}(\mathbf{I}{+}\mathbf{A})$ matrices, both predicted from the input histogram by a shared transformer, and show across per-sensor and mixed-sensor RAW benchmarks that it outperforms previous task-oriented ISPs at near-constant inference cost.

\paragraph{Limitation}
Real mixed-sensor RAW data with balanced labels remains scarce, so our experiments on natively heterogeneous captures remain challenging to conduct at large scale. Developing such a label-balanced, multi-sensor RAW detection dataset would be a valuable direction for future research.

\appendix

\section{Simulated RAW Augmentation}
\label{sec:raw_aug}

To improve the generalization of our detection model across heterogeneous imaging conditions, we construct a physics-based RAW image simulation pipeline that synthesizes plausible sensor outputs under diverse camera spectral sensitivities, illuminants, exposures, and sensor characteristics. All operations are performed in linear RAW space after black-level subtraction, and the output is a simulated linear RAW image normalized to $[0, 1]$. \Cref{tab:hybrid_dataset} reports detection results on our synthesized dataset, which is built from PASCALRAW~\cite{omid2017toward}, LOD~\cite{hong2021crafting}, and ROD~\cite{xu2023toward}. We exclude AODRaw~\cite{li2025towards} because it already contains diverse real-world degradations such as adverse weather and challenging illumination.

\begin{figure}[!ht]
    \centering
    \includegraphics[width=\textwidth]{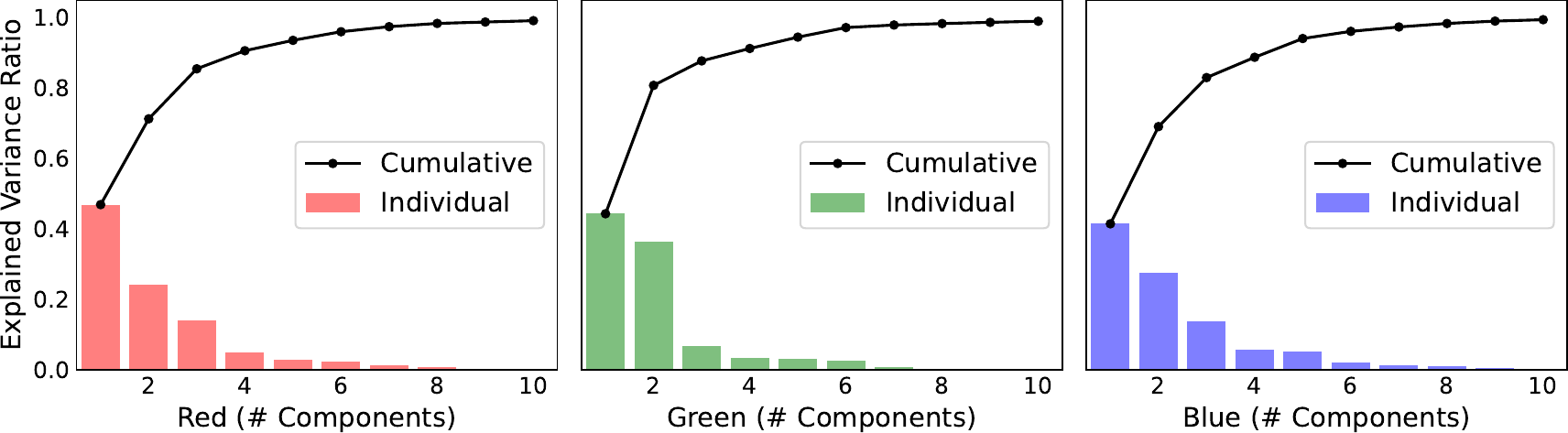}
    \caption{Per-channel PCA explained variance for the 28-camera spectral sensitivity database.}
    \label{fig:sup_pca}
\end{figure}

\paragraph{Camera Spectral Sensitivity Database.}
We use the 28-camera spectral sensitivity database of \cite{jiang2013camspec}, which provides per-channel sensitivity curves $S_c^{(i)}(\lambda)$ for $c \in \{R, G, B\}$ sampled at 33 wavelengths from 400\,nm to 720\,nm in 10\,nm increments. Each curve is peak-normalized to unity. Following \cite{jiang2013camspec}, we observe that the space of real camera sensitivities is well approximated by a low-dimensional linear subspace, and fitting a two-component PCA independently for each channel explains most of the inter-camera variance as shown in \Cref{fig:sup_pca}.

\paragraph{Offline Precomputation.}
Let $\mathbf{X}^c \in \mathbb{R}^{28 \times 33}$ denote the matrix of peak-normalized sensitivity curves for channel $c$. We fit PCA with $k=2$ components as:
\begin{equation}
    \mathbf{X}^c = \mathbf{1}(\boldsymbol{\mu}^c)^\top +
    \mathbf{1}(\boldsymbol{\sigma}^c)^\top \odot
    \mathbf{Z}^c \mathbf{V}^{c\top} + \mathbf{E}^c,
\end{equation}
where $\mathbf{V}^c \in \mathbb{R}^{33 \times 2}$ are the principal components and $\mathbf{Z}^c \in \mathbb{R}^{28 \times 2}$ are the corresponding coefficients. We record the per-component empirical mean $\bar{\mathbf{z}}^c$, standard deviation $\hat{\sigma}^c$, and extrema $(\mathbf{z}_{\min}^c,\, \mathbf{z}_{\max}^c)$ from the 28 real cameras.

\begin{figure}[!tp]
    \centering
    \includegraphics[width=\textwidth]{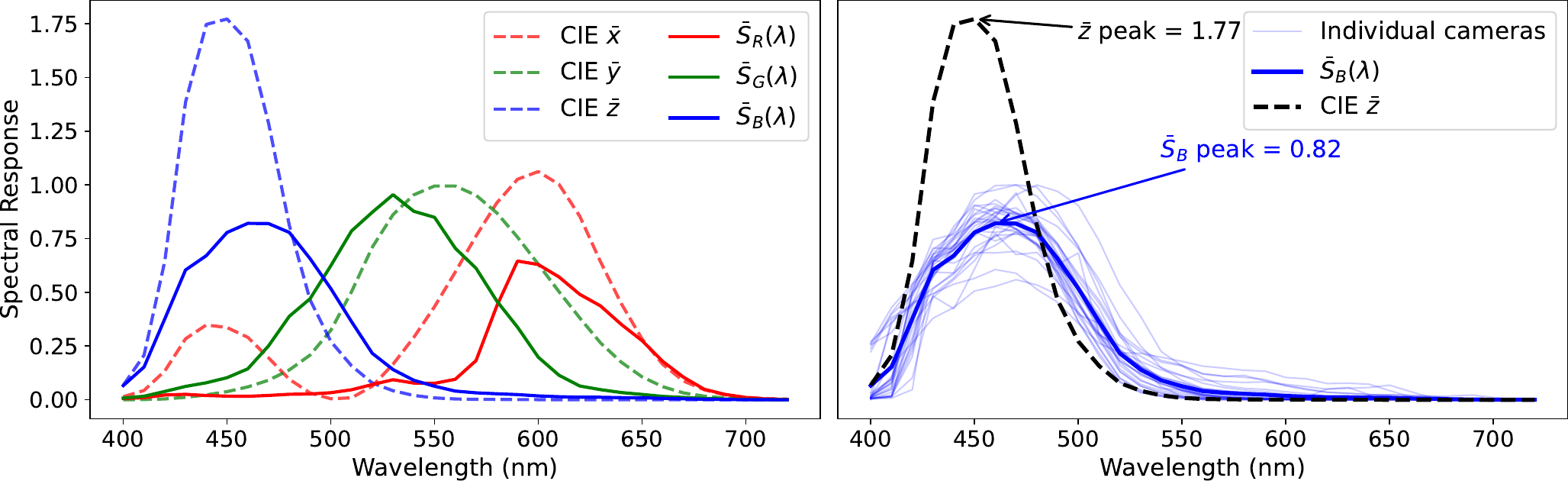}
    \caption{Comparison of the CIE 1931 2\textdegree\ standard observer CMFs (dashed) and the 28-camera mean spectral sensitivity $\bar{S}(\lambda)$ (solid) used as the source camera proxy. The CIE $\bar{z}$ function peaks at 1.77, far above the typical blue-channel response of real cameras (${\approx}0.82$), motivating our use of $\bar{S}(\lambda)$ to avoid compressing the blue-channel dynamic range in the cross-camera transform.}
    \label{fig:sup_cie_cmf}
\end{figure}

We also precompute the ColorChecker response of a surrogate source camera. Since the true spectral sensitivity of the source camera is generally unavailable, we approximate it by the mean sensitivity across all 28 cameras:
\begin{equation}
    \bar{S}(\lambda) = \frac{1}{28} \sum_{i=1}^{28} S^{(i)}(\lambda).
\end{equation}
This choice is preferred over the CIE 1931 2\textdegree\ standard observer~\cite{cie1931} as the source proxy. As illustrated in~\Cref{fig:sup_cie_cmf}, the CIE $\bar{z}$ function peaks at 1.77, well above the typical blue-channel response of real cameras (${\approx}0.82$), which would artificially suppress the blue-channel dynamic range in the estimated cross-camera matrix. Under D65 illumination~\cite{cied65}, the ColorChecker response of the source proxy is
\begin{equation}
    \mathbf{P}_A = \bigl(\mathbf{R} \odot \mathbf{d}_{65}\mathbf{1}^\top\bigr)^\top
    \bar{\mathbf{S}} \cdot \Delta\lambda \in \mathbb{R}^{24 \times 3},
\end{equation}
where $\mathbf{R} \in \mathbb{R}^{33 \times 24}$ contains the Macbeth ColorChecker 2005 reflectance spectra~\cite{mccamy1976color}, $\mathbf{d}_{65} \in \mathbb{R}^{33}$ is the D65 spectral power distribution, and $\Delta\lambda = 10\,\text{nm}$. $\mathbf{P}_A$ is then normalized so that the green column equals unity.

\paragraph{Exposure Sampling.}
A global exposure gain $\alpha$ is sampled log-uniformly in stop units:
\begin{equation}
    u \sim \mathcal{U}(u_{\min},\, u_{\max}), \qquad \alpha = 2^u,
\end{equation}
with $u_{\min} = -3$ and $u_{\max} = 3$, spanning a six-stop dynamic range. This log-uniform prior reflects the fact that perceptually equal exposure steps correspond to multiplicative gains~\cite{reinhard2010hdr}.

\begin{figure}[!ht]
    \centering
    \includegraphics[width=\textwidth]{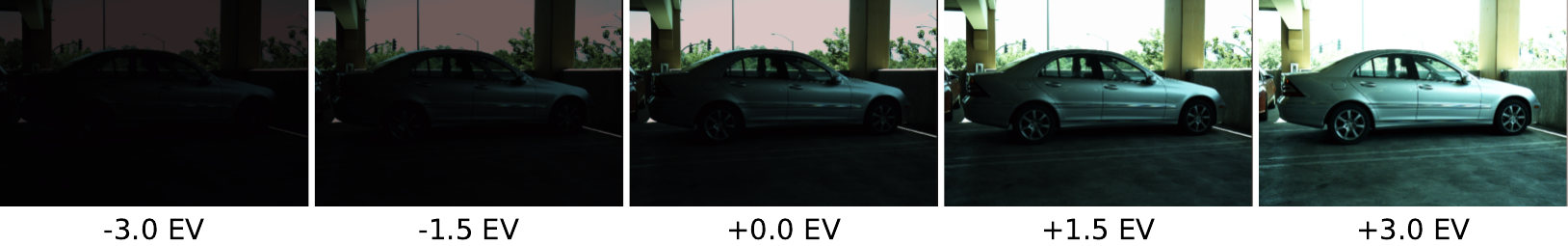}
    \caption{Visualization of RAW images synthesized using different exposure values (EVs).}
    \label{fig:sup_exposure}
\end{figure}

\begin{figure}[!ht]
    \centering
    \includegraphics[width=0.5\textwidth]{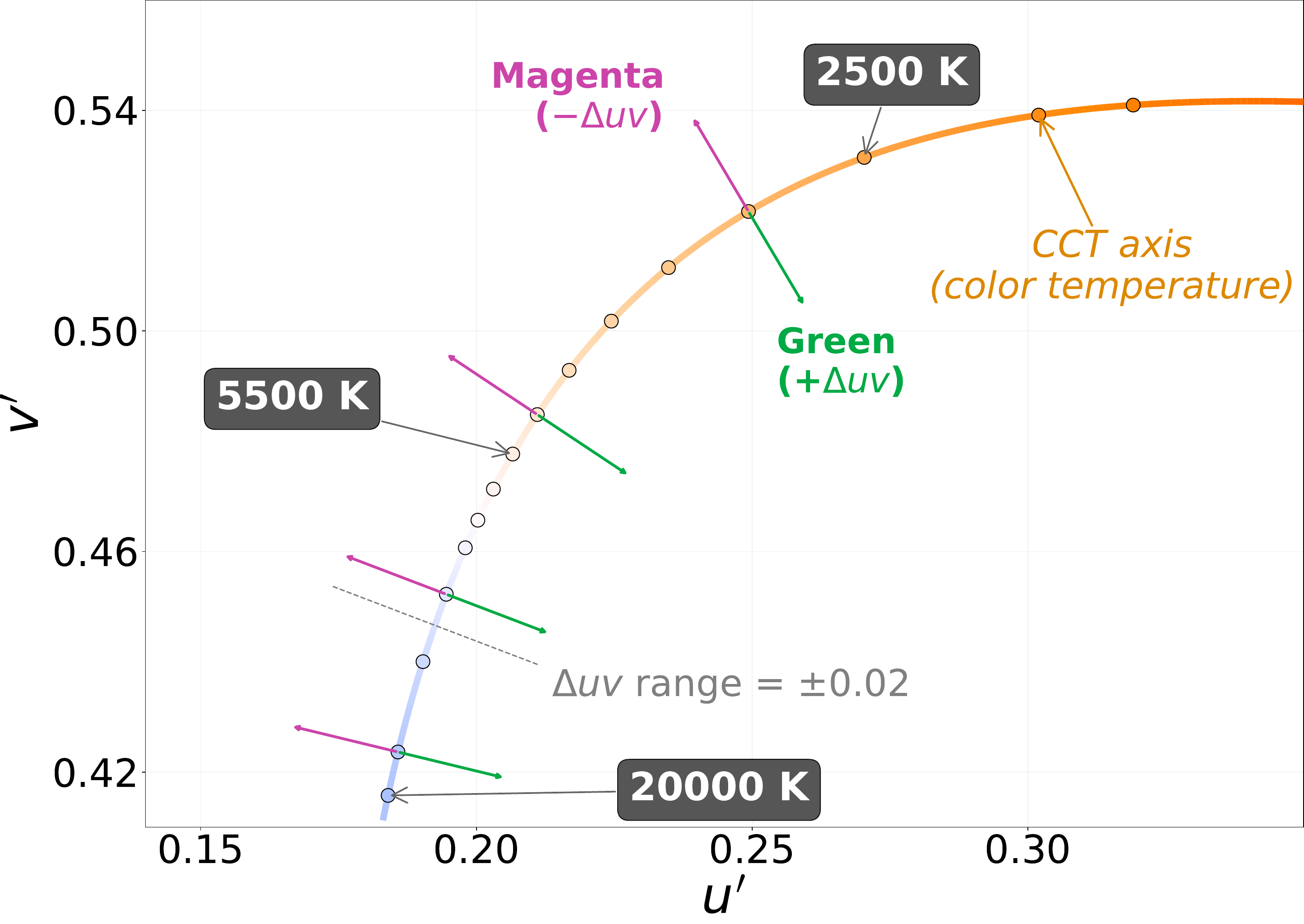}
    \caption{Visualization of Planckian Locus on CIE 1976 $u'v'$ chromaticity diagram.}
    \label{fig:sup_planckian_locus}
\end{figure}

\begin{figure}[!ht]
    \centering
    \begin{minipage}[t]{0.49\linewidth}
        \centering
        \includegraphics[width=\linewidth]{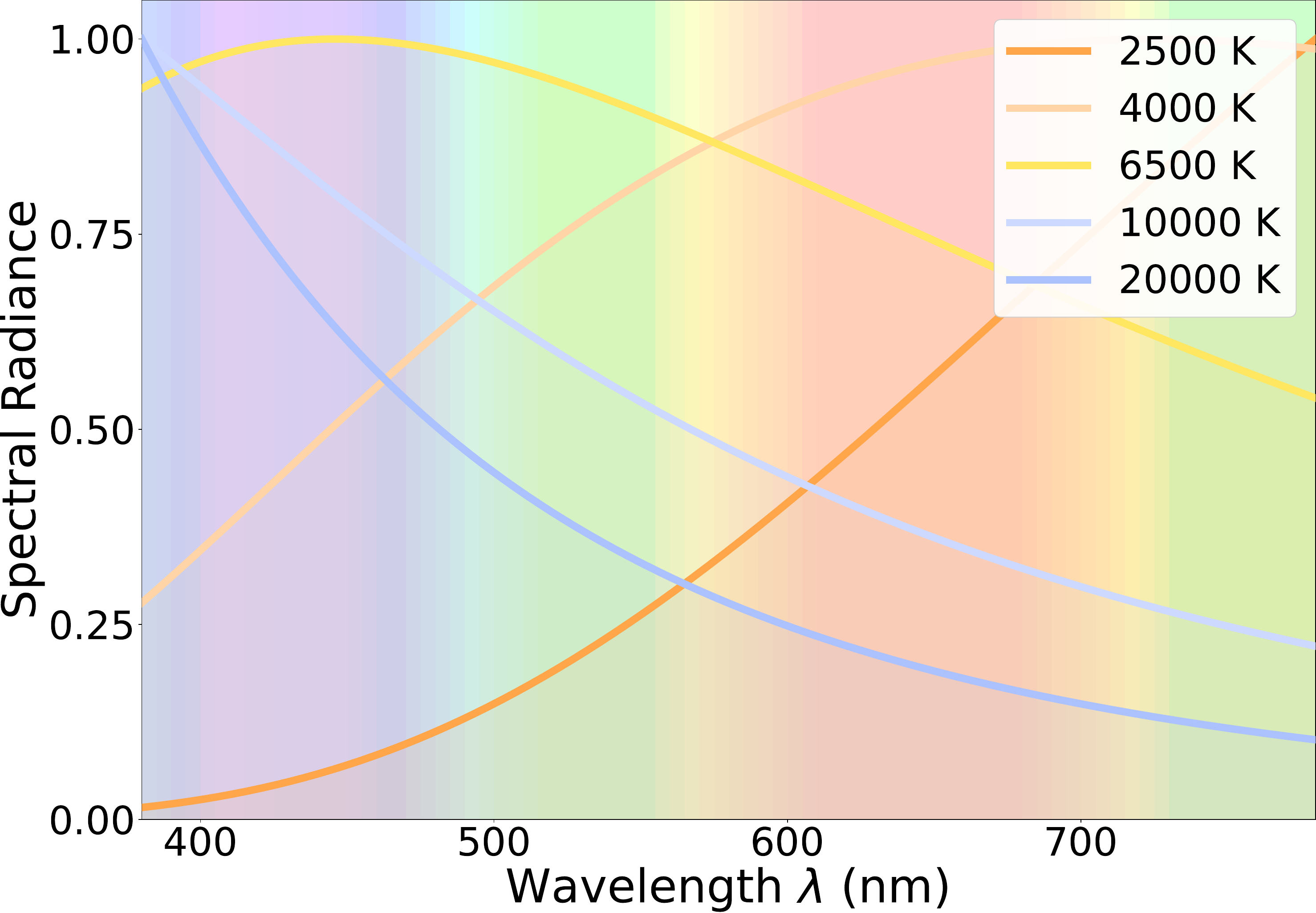}
        \caption{Spectral power distributions $B(\lambda, T)$ of black-body radiators at representative color temperatures spanning our sampling range from 2{,}500\,K to 20{,}000\,K, computed via Planck's law and
        peak-normalized to unity.}
        \label{fig:sup_blackbody_spectra}
    \end{minipage}
    \hfill
    \begin{minipage}[t]{0.49\linewidth}
        \centering
        \includegraphics[width=\linewidth]{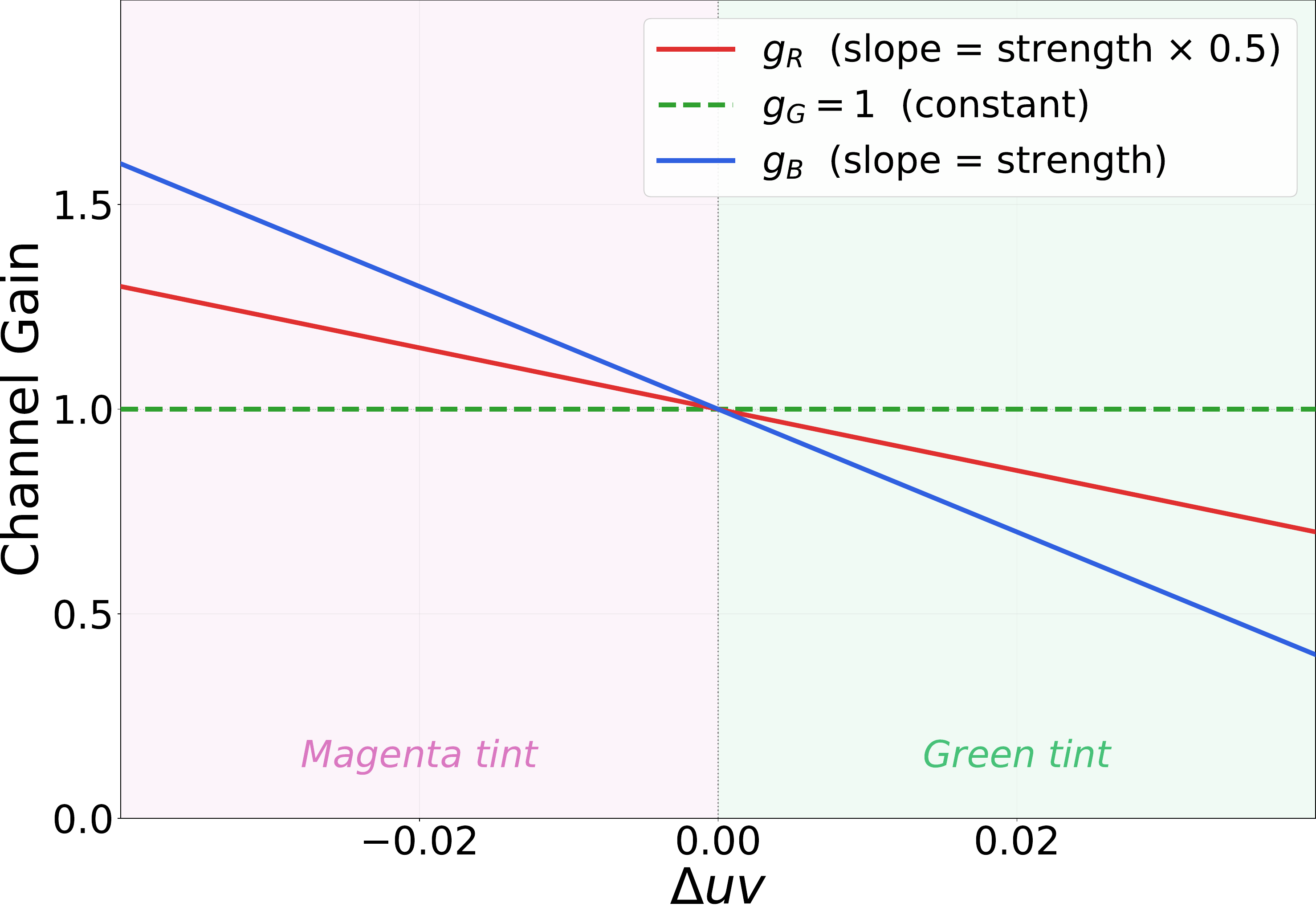}
        \caption{Diagonal RGB gains $\mathbf{g}_{\text{tint}}$ as a function of tint offset $\Delta uv$. The R channel receives half the correction of the B channel, preventing over-reddening under strong magenta tints while keeping the G gain fixed at unity.}
        \label{fig:sup_tint_gain}
    \end{minipage}
\end{figure}

\paragraph{Color Temperature and Tint Sampling.}
We sample the illuminant color temperature in the mired scale~\cite{priest1933proposed}, also known as the micro-reciprocal degree scale. This parameterization is commonly used for color-temperature shifts because equal steps in mired produce more uniform perceptual changes than equal steps in Kelvin, especially along the Planckian
locus~\cite{wyszecki2000color}. As shown in \Cref{fig:sup_planckian_locus}, we uniformly sample the mired value and convert it to the corresponding color temperature as
\begin{equation}
    m \sim \mathcal{U}(m_{\min},\, m_{\max}), \qquad T = \frac{10^6}{m}\,\text{K},
\end{equation}
with $m_{\min} = 50$ corresponding to $T = 20{,}000\,\text{K}$ for extreme daylight and $m_{\max} = 400$ corresponding to $T = 2{,}500\,\text{K}$ for tungsten illumination. Given the sampled temperature, we compute the
corresponding spectral power distribution using Planck's law of black-body radiation~\cite{planck1901},
\begin{equation}
    B(\lambda, T) =
    \frac{2\pi h c^2}{\lambda^5
    \left[\exp\!\left(\dfrac{hc}{k_B T \lambda}\right) - 1\right]},
\end{equation}
where $h$ is the Planck constant, $c$ the speed of light, and $k_B$ the Boltzmann constant. The resulting spectrum $I_T(\lambda)$ is peak-normalized to unity. \Cref{fig:sup_blackbody_spectra} illustrates the spectral shift from warm to cool illuminants across the sampled range.

\begin{figure}[!ht]
    \centering
    \includegraphics[width=\textwidth]{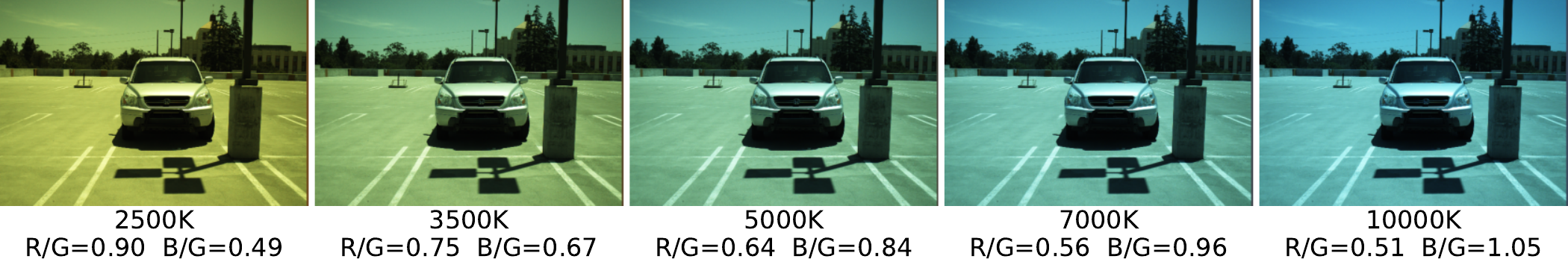}
    \caption{Visualization of color temperature variation on a real RAW image, simulated across the full sampled range from warm tungsten illumination at 2{,}500\,K to extreme cool daylight at 20{,}000\,K.}
    \label{fig:sup_temperature}
\end{figure}

\begin{figure}[!ht]
    \centering
    \includegraphics[width=\textwidth]{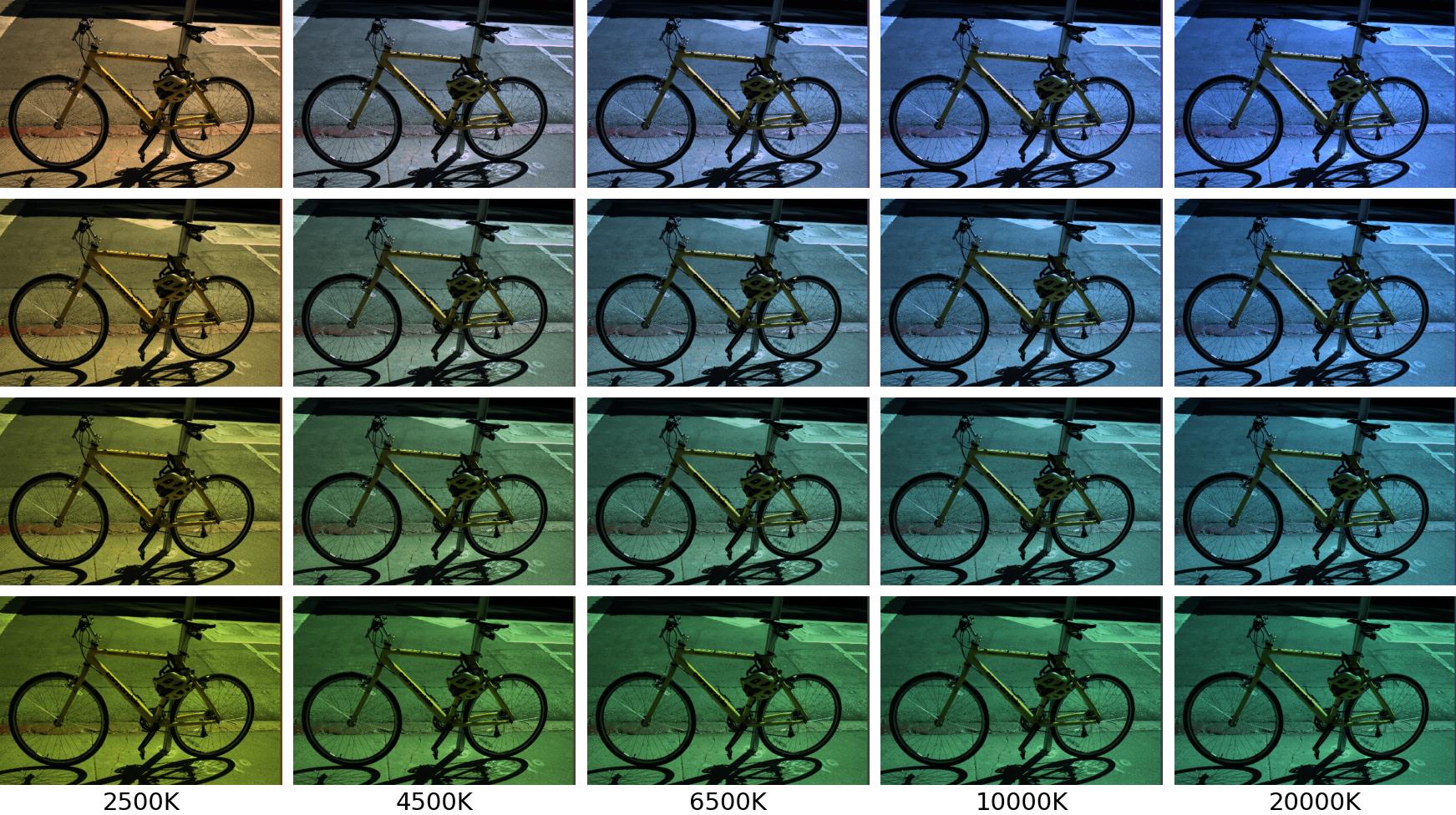}
    \caption{Visualization of temperature and tint variation on a real RAW image, simulated across the full sampled range from strong magenta at $\Delta uv = -0.04$ to strong green at $\Delta uv = +0.04$ along the green--magenta axis of the CIE $u'v'$ chromaticity diagram.}
    \label{fig:sup_tint}
\end{figure}

To account for real-world deviations from the ideal Planckian locus, such as those produced by fluorescent or LED sources, we additionally sample a tint offset $\Delta uv \sim \mathcal{U}(-0.04, 0.04)$ along the green--magenta axis of the CIE $u'v'$ uniform chromaticity diagram~\cite{cie1976}. Rather than applying this offset via spectral modification, which requires inverting an overdetermined $33 \times 3$ system and yields numerically unstable results, we represent tint as a diagonal RGB gain applied after the cross-camera transform,
\begin{equation}
    \mathbf{g}_{\text{tint}} =
    \bigl[\,\max(1 - 0.5 \cdot s \cdot \Delta uv,\; 0.1),\;\;
            1.0,\;\;
            \max(1 - s \cdot \Delta uv,\; 0.1)\,\bigr]^\top,
\end{equation}
where $s = 15.0$ is a fixed scaling constant. The asymmetric weighting, where R receives half the correction of B, prevents over-reddening under strong magenta tints, as reflected in the gain curves shown in \Cref{fig:sup_tint_gain}. The perceptual effects of color temperature and tint variation on a real RAW image are illustrated in \Cref{fig:sup_temperature,fig:sup_tint}, where the simulated illuminant shifts from warm tungsten to cool daylight and from green to magenta across the full sampled ranges.

\begin{figure}[!ht]
    \centering
    \includegraphics[width=\textwidth]{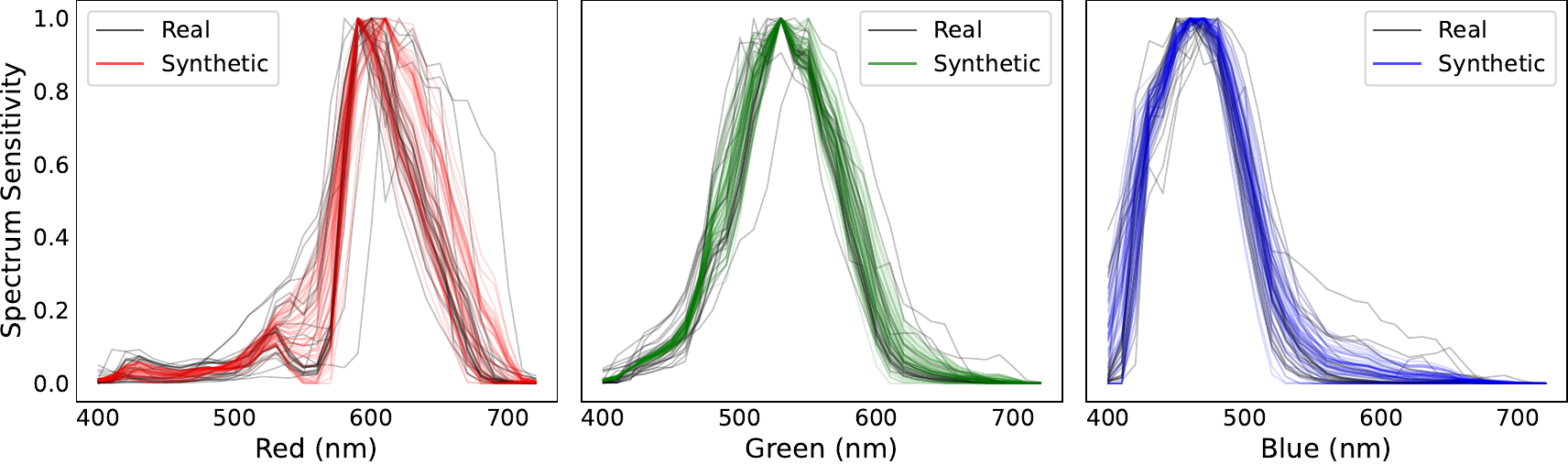}
    \caption{Visualization of spectral sensitivity of real cameras and our synthesized responses.}
    \label{fig:sup_spectrum_response}
\end{figure}

\paragraph{Target Camera Spectral Response Sampling.}
A virtual target camera sensitivity $S_B(\lambda) \in \mathbb{R}^{33 \times 3}$ is sampled from the learned low-dimensional manifold. For each channel $c$, we draw PCA coefficients from a truncated Gaussian:
\begin{equation}
    \boldsymbol{\zeta}^c \sim
    \mathcal{N}\bigl(\bar{\mathbf{z}}^c,\; \mathrm{diag}(\hat{\sigma}^c)^2\bigr),
    \quad \text{clipped to}\;\; [\mathbf{z}_{\min}^c, \mathbf{z}_{\max}^c].
\end{equation}
The sensitivity curve is then reconstructed and constrained to be physically realizable:
\begin{equation}
    \tilde{S}^c(\lambda) = \boldsymbol{\sigma}^c \odot \mathbf{V}^c \boldsymbol{\zeta}^c + \boldsymbol{\mu}^c,
    \qquad
    S_B^c(\lambda) = \frac{\max(\tilde{S}^c(\lambda), 0)}
    {\max_\lambda \tilde{S}^c(\lambda) + \epsilon}.
\end{equation}
The Gaussian prior concentrates samples near the center of the observed
camera distribution, while the clipping to $[\mathbf{z}_{\min}^c,
\mathbf{z}_{\max}^c]$ ensures the reconstructed curves remain within
the convex hull spanned by real cameras. As shown in ~\Cref{fig:sup_spectrum_response}, the synthetic sensitivities faithfully reproduce the shape and spread of the diverse real cameras across all three channels, while providing substantially denser coverage of the plausible sensitivity space.

\paragraph{Cross-camera Color Transform.}
We compute a $3\times 3$ linear matrix $\mathbf{M}_{A \to B,T}$ that jointly accounts for the change in camera spectral sensitivity and the change in illuminant from D65 to $I_T$. The ColorChecker response of the target camera under illuminant $I_T$ is
\begin{equation}
    \mathbf{P}_B = \bigl(\mathbf{R} \odot I_T \mathbf{1}^\top\bigr)^\top
    \mathbf{S}_B \cdot \Delta\lambda \in \mathbb{R}^{24 \times 3},
\end{equation}
normalized so that the green column equals unity. The cross-camera matrix is estimated by least-squares regression over the 24 ColorChecker patches:
\begin{equation}
    \mathbf{M}_{A \to B,T} = \operatorname*{arg\,min}_{\mathbf{M}}
    \|\mathbf{P}_A \mathbf{M}^\top - \mathbf{P}_B\|_F^2.
\end{equation}
This bridge approach, which uses a reflectance chart to relate two cameras under different illuminants, follows the methodology established in camera color characterization~\cite{barnard2002comparison,finlayson2015color}.

\begin{figure}[!tp]
    \centering
    \begin{minipage}[t]{0.49\linewidth}
        \centering
        \includegraphics[width=\linewidth]{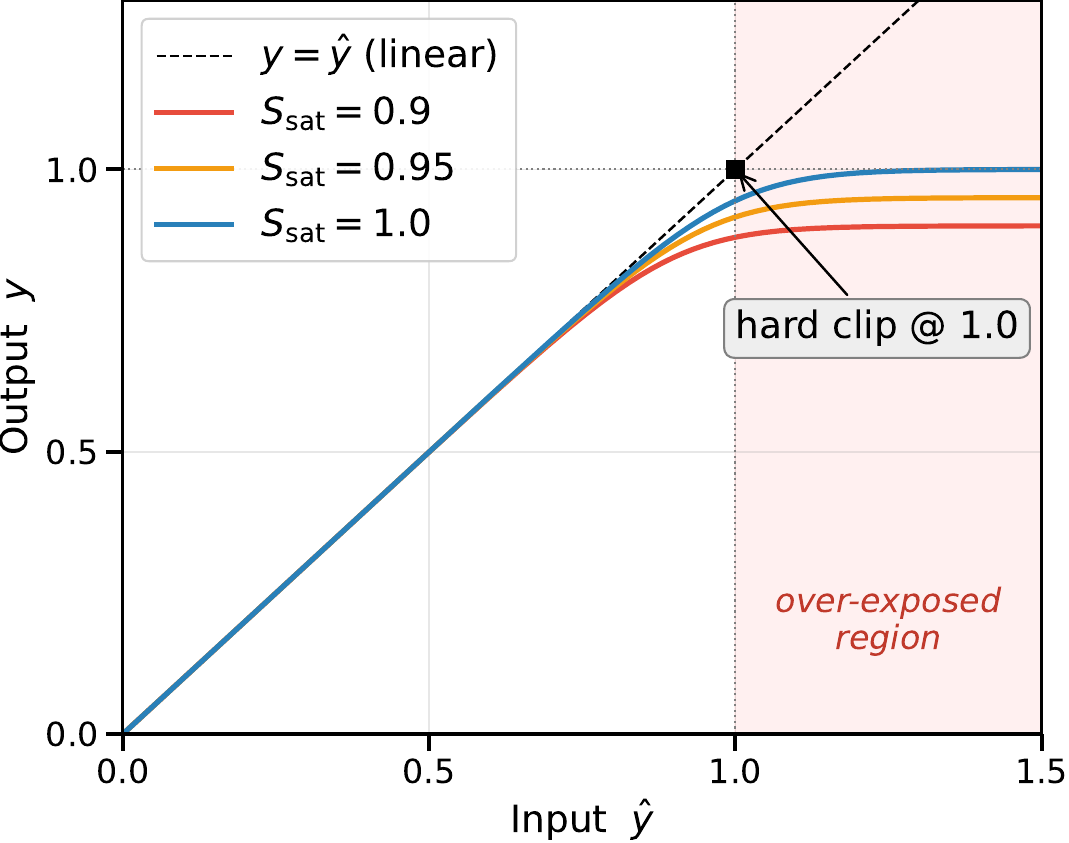}
        \caption{Illustration of the sensor saturation model. For inputs below the saturation threshold $S_\mathrm{sat}$, the response follows a smooth compressive curve; beyond this point, the response asymptotically approaches $S_\mathrm{sat}$ through the $\tanh$ roll-off. Lower $S_\mathrm{sat}$ values produce stronger compression and an earlier onset of saturation.}
    \label{fig:sup_cutoff}
    \end{minipage}
    \hfill
    \begin{minipage}[t]{0.49\linewidth}
        \centering
        \includegraphics[width=\linewidth]{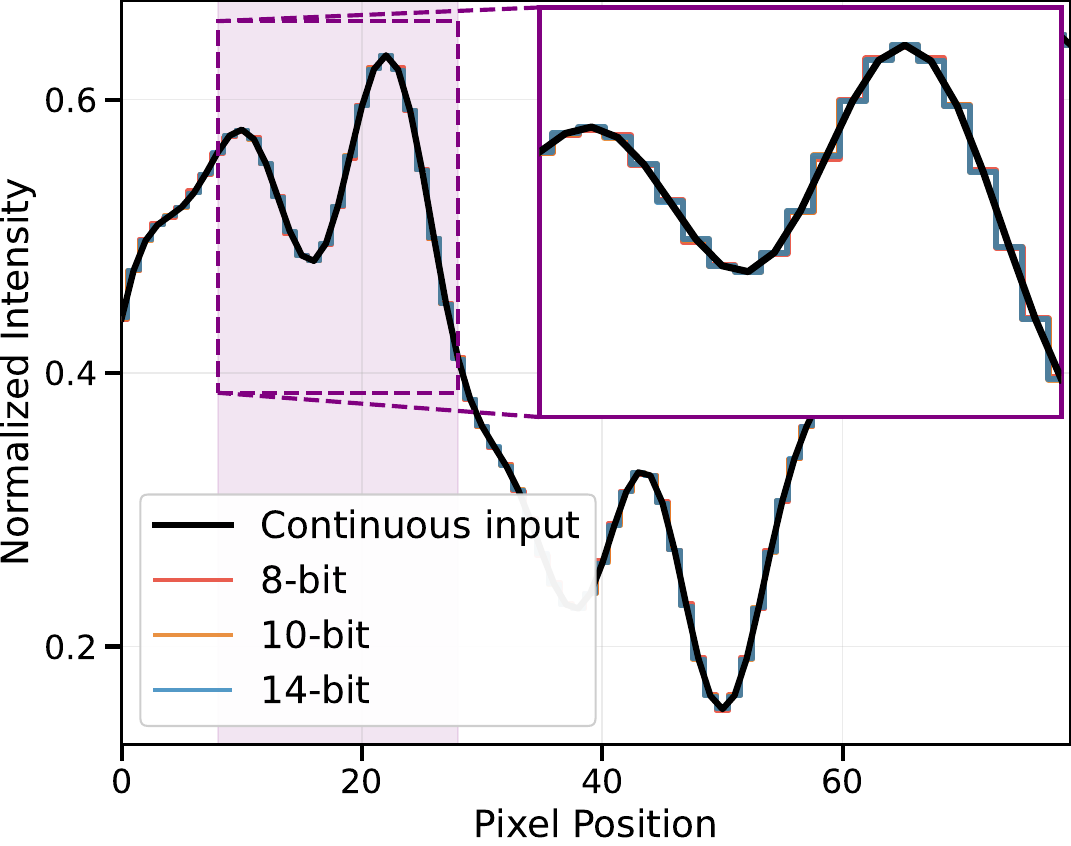}
        \caption{Effect of ADC bit depth on a representative 1D intensity profile. Coarser quantization (8-bit) introduces visible staircase artifacts relative to the continuous reference, while finer quantization (10-bit and 14-bit) progressively recovers the smooth signal shape.}
    \label{fig:sup_quant}
    \end{minipage}
\end{figure}

\paragraph{Overall Transformation.}
Having sampled the exposure $\alpha$, the cross-camera matrix $\mathbf{M}_{A \to B,T}$, and the tint gain $\mathbf{g}_{\mathrm{tint}}$, we further model sensor non-idealities via a channel-crosstalk matrix $\boldsymbol{\varepsilon}$, with off-diagonal entries drawn from $\mathcal{U}(-0.05, 0.05)$ and diagonal set to zero, and per-channel black-level offsets $\boldsymbol{\beta} \sim \mathcal{U}(-0.02, 0.02)^3$. Before applying the transform, we remove the inherent green bias of Bayer RAW images via channel-wise white balance:
\begin{equation}
    \mathbf{g}_{\mathrm{wb}} = \bar{c}_G \big/ \bar{\mathbf{c}},
    \qquad
    \mathbf{x}_{\mathrm{bal}} = \mathbf{x} \odot \mathbf{g}_{\mathrm{wb}},
\end{equation}
where $\bar{\mathbf{c}}$ denotes per-channel spatial means. The full transform matrix is then assembled as
\begin{equation}
    \mathbf{M}_{\mathrm{full}} = \alpha \bigl(
    \mathrm{diag}(\mathbf{g}_{\mathrm{tint}})\,
    \mathbf{M}_{A \to B,T} + \boldsymbol{\varepsilon}\bigr),
\end{equation}
and applied pixel-wise:
\begin{equation}
    \mathbf{y} = \mathbf{M}_{\mathrm{full}}\, \mathbf{x}_{\mathrm{bal}}
    + \boldsymbol{\beta}.
\end{equation}
The white balance is then reversed using target camera B's own gray-world multipliers $\mathbf{g}_{\mathrm{wb}}^B = r_{B,G}/\mathbf{r}_B$ with $\mathbf{r}_B = \mathbf{S}_B^\top I_T \cdot \Delta\lambda$, rather than the source camera's $\mathbf{g}_{\mathrm{wb}}$, so that the output retains target camera B's natural green dominance under illuminant $I_T$ rather than re-imprinting the source camera's spectral bias:
\begin{equation}
    \hat{\mathbf{y}} = \mathbf{y} \oslash \mathbf{g}_{\mathrm{wb}}^B.
\end{equation}

\paragraph{Highlight Roll-off and Quantization.}
To simulate the soft saturation behavior of real sensors approaching full-well capacity~\cite{healey1994radiometric}, we apply a $\tanh$-based roll-off:
\begin{equation}
    \hat{\mathbf{y}}' = S_{\mathrm{sat}}\,
    \tanh\!\left(\frac{\hat{\mathbf{y}}}{S_{\mathrm{sat}}}\right),
    \quad S_{\mathrm{sat}} \sim \mathcal{U}(0.9, 1.0),
\end{equation}
where the $\tanh$ function provides a smooth asymptotic upper bound that approaches but never exceeds $S_{\mathrm{sat}}$, while any sub-zero values arising from the black-level offset $\boldsymbol{\beta}$ are clipped to zero, as illustrated in Figure~\ref{fig:sup_cutoff}. Finally, the output is quantized to a bit depth $b \in \{10, 12, 14, 16\}$ chosen uniformly at random:

\begin{equation}
    \tilde{\mathbf{y}} = \frac{1}{2^b - 1}
    \left\lfloor \frac{\hat{\mathbf{y}}'}{S_{\mathrm{sat}}} (2^b - 1) + 0.5 \right\rfloor,
\end{equation}
where the floor operation is applied element-wise. The effect of bit depth on signal fidelity is shown in \Cref{fig:sup_quant}. Since every operation is a pointwise color transformation with no spatial warping, ground-truth bounding box annotations remain valid without modification.

\section{Additional Related Works}

\paragraph{RAW Imaging Pipelines for High-Level Vision.}
Early reconfigurable imaging pipelines~\cite{buckler2017reconfiguring} demonstrate that bypassing most ISP stages preserves the accuracy of CNN-based vision while reducing sensor power, motivating the use of RAW data as a direct input to perception networks. Subsequent works learn end-to-end pipelines that fuse a lightweight neural front end with a downstream classifier or detector~\cite{diamond2021dirty,yoshimura2023dynamicisp,cui2025raw}. Beyond accuracy, RAW data also provides natural robustness signals, with~\cite{zhang2022all} mapping RGB inputs through a learned RAW-domain ISP to defend against adversarial perturbations across classification, segmentation, and detection.

\paragraph{Low-Light RAW Object Detection.}
Low-light scenarios drive a substantial body of RAW-specific detection research. \cite{sasagawa2020yolo} merges a learned dark-image enhancer with YOLO via glue layers and knowledge distillation. \cite{cui2021multitask} introduces a multitask auto-encoding-transformation framework that disentangles ISP and illumination factors from object features. \cite{hashmi2023featenhancer} learns hierarchical feature enhancers trained jointly with the detector, while \cite{hong2024you} extracts illumination-invariant kernels grounded in the Lambertian image-formation model. Beyond static images, \cite{wang2024multi} addresses multi-object tracking on dark RAW video with adaptive low-pass downsampling and degradation-suppression learning.

\paragraph{RAW Segmentation and Other High-Level Tasks.}
Beyond object detection, RAW data has been studied for additional high-level tasks. \cite{chen2023instance} performs instance segmentation directly on RAW low-light captures with feature-noise suppression, smooth convolutions, and a paired LIS benchmark. \cite{sun2023adaptive} formulates shadow detection on RAW HDR data through a many-to-many adaptive illumination mapping that preserves multi-scale contrast often lost after ISP tone compression. These efforts show that the linear photon evidence retained in RAW captures benefits dense prediction tasks well beyond bounding-box detection.

\paragraph{sRGB-to-RAW Reconstruction.}
A complementary research thread reverses the camera pipeline to synthesize realistic RAW from sRGB inputs, enabling RAW-side training without paired sensor captures. The unprocessing framework~\cite{brooks2019unprocessing} inverts each ISP step in closed form to render large RGB datasets back to the sensor domain for raw denoiser training at scale. CycleISP~\cite{zamir2020cycleisp} introduces a cycle-consistent learned ISP and inverse-ISP pair that produces realistic noisy RAW for image restoration. Invertible ISP~\cite{xing2021invertible} replaces the camera pipeline with a normalizing flow whose forward pass renders sRGB and inverse pass recovers near-perfect RAW with no metadata. A parallel line stores compact metadata alongside sRGB to anchor accurate per-pixel reconstruction, evolving from raw-reconstruction-aware sRGB compressors~\cite{punnappurath2019learning} and spatially-aware sample pairs~\cite{punnappurath2021spatially} to content-aware metadata~\cite{nam2022learning} and end-to-end learned latent metadata~\cite{wang2023raw,wang2024beyond}, with a recent extension to video sequences~\cite{zhang2024leveraging}.

\paragraph{Synthetic RAW for Downstream Tasks.}
Synthesized RAW data has been widely used to scale learning beyond paired captures. Physics-based noise modeling~\cite{wei2020physics} calibrates sensor shot, read, row, and quantization noise to match real low-light RAW statistics. Neural Camera Simulators~\cite{ouyang2021neural} learn a controllable RAW synthesizer conditioned on capture parameters such as exposure and ISO. Day-to-night RAW synthesis~\cite{punnappurath2022day} relights daytime RAW captures into nighttime variants for training neural ISPs. ReRAW~\cite{berdan2025reraw} reconstructs sensor-specific RAW from large RGB datasets via stratified sampling for efficient detector pre-training on edge sensors. These inverse-ISP and synthetic-RAW designs are complementary in spirit to our physics-based simulation, while our adapter focuses on a richer global-local tonal representation conditioned on per-image RAW statistics.

\paragraph{RAW for Scientific Imaging.}
Beyond consumer photography, RAW captures play a central role in scientific instruments where every photon carries quantitative meaning. In fluorescence microscopy, content-aware image restoration~\cite{weigert2018content} pioneered the use of deep networks on raw microscope volumes for denoising and isotropy recovery, followed by label-free virtual fluorescence prediction from transmitted light~\cite{ounkomol2018label}, cross-modality super-resolution~\cite{wang2019deep}, accelerated single-molecule localization microscopy~\cite{ouyang2018deep,nehme2020deepstorm3d,speiser2021deep}, Fourier-channel attention super-resolution~\cite{qiao2021evaluation}, and three-dimensional residual channel attention denoising~\cite{chen2021three}, all surveyed in~\cite{belthangady2019applications}. In astronomy, deep networks operate directly on raw CCD frames for galaxy-scale gravitational lensing analysis~\cite{hezaveh2017fast}, self-supervised denoising of ground-based images~\cite{liu2025astronomical}, and self-supervised spatiotemporal denoising that pushes deeper detection limits~\cite{guo2026deeper}. Closer to our setting, physics-based CCD noise modeling tailored to deep-sky observation~\cite{liu2026denoising} and conservative flow-matching for ground-to-space astronomical super-resolution~\cite{liu2026fluxflow} demonstrate that the linear photon evidence retained in scientific RAW captures supports both restoration and downstream measurement. These developments underline that sensor-agnostic RAW modeling, the focus of our work, is increasingly relevant across imaging domains far beyond ordinary RGB photography.

\paragraph{Broader sRGB Image Processing.}
In the 2D image domain, recent work spans data-centric denoising~\cite{chang2026beyond}, training-free ensembling~\cite{chang2026training}, and adverse-condition image enhancement~\cite{ge2026clip}. In remote sensing, infrared image super-resolution has been explored~\cite{ge2026dual}. In 3D vision, related efforts cover neural field rendering with media interaction~\cite{liu2025i2}, structural-prior multi-view reconstruction~\cite{liu2025mg}, scene reconstruction under adverse weather~\cite{liu2025deraings}, and physically-degraded multi-view benchmarks~\cite{liu2025realx3d,liu2026ntire}. These pipelines operate on three-channel or multi-view RGB inputs rather than RAW image domains.

\section{Additional Experimental Details and Results}
\label{sec:appendix_experiment}

This section summarizes the data curation and shared evaluation protocol used for MultiRAW in our mixed-sensor experiments, and provides more visualizations for the RAW object detection experiments in \Cref{sec:experiment}, supplementary qualitative results from the LOD ablation study in \Cref{sec:ablation}, and additional experiments on RAW image segmentation using the ADE20K dataset~\cite{zhou2019semantic}.

\subsection{MultiRAW Dataset}
\label{sec:multiraw_protocol}
MultiRAW~\cite{li2024efficient} is a multi-device RAW detection benchmark of 7,208 images captured by the four sensors \texttt{ASI\_294MCPro}, \texttt{Huawei\_P30Pro}, \texttt{iPhone\_XSMax}, and \texttt{OnePlus\_5T}, spanning heterogeneous CFA patterns, resolutions, and bit depths ranging from 10 to 24 bits, and accompanied by per-sensor ISP renderings and XML annotations.

However, the MultiRAW dataset~\cite{li2024efficient} exhibits severe cross-sensor class imbalance. For instance, the sensor \texttt{oneplus\_5t} appears only in the test split. The 11-class taxonomy is also long-tailed and device-coupled, and contains ambiguous overlapping categories such as the merged \texttt{bicycle}, \texttt{motorcycle}, and \texttt{tricycle}. We further observe label misalignment on \texttt{huawei\_p30pro}, where the ground-truth annotations are shifted from the actual objects in the RAW images.

To address these problems, we build a unified preprocessing pipeline that re-associates RAW images with metadata and corrects label misalignment. Specifically, we keep the original data split intact and select the top-five classes shared across all four sensors, namely \texttt{car}, \texttt{traffic-sign}, \texttt{traffic-light}, \texttt{person}, and \texttt{bicycle}, and rebalance categories by randomly sampling each class. This yields a class-balanced MultiRAW subset of $3{,}500$ training images and 2{,}064 test images, on which all baselines are trained and evaluated as shown in \Cref{tab:hybrid_dataset}.

\subsection{Additional Qualitative Results for RAW Object Detection}
\label{app:additional_detection}

\begin{figure}[!ht]
    \centering
    \includegraphics[width=\textwidth]{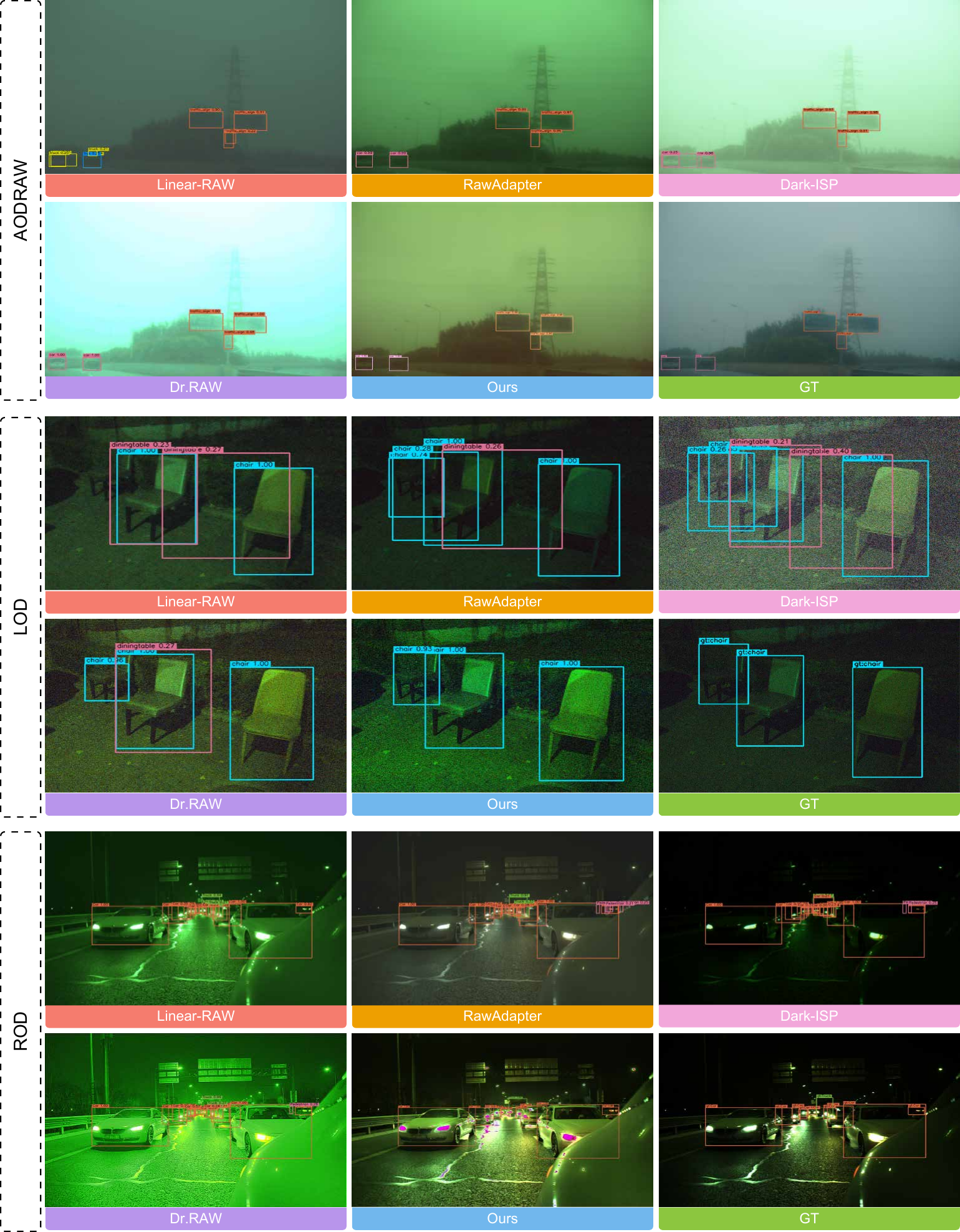}
    \caption{Additional qualitative object detection results on AODRAW~\cite{li2025towards}, LOD~\cite{hong2021crafting}, and ROD~\cite{xu2023toward}. The examples highlight adverse weather, severe low-light noise, and simultaneous highlights and shadows, under which our method produces cleaner detections and more stable localization than prior task-oriented ISP baselines.}
    \label{fig:sup_detection}
\end{figure}

We show qualitative results on three RAW detection benchmarks that together
span the main difficulties of in-the-wild RAW imaging.
AODRAW ~\cite{li2025towards} covers diverse indoor and outdoor scenes under
challenging illumination and adverse weather, including rain, haze, low
light, and reduced contrast. LOD ~\cite{hong2021crafting} focuses on
extremely low-light indoor scenes where signal-to-noise ratio is the
dominant bottleneck. ROD~\cite{xu2023toward} consists of automotive driving
scenes captured in 24-bit HDR RAW, exposing methods to wide dynamic range
and strong highlight or shadow regions typical of day and night traffic.
\Cref{fig:sup_detection} visualizes predictions from our method alongside
task-oriented ISP baselines on representative images from each dataset.
Across these cases, our method yields cleaner boxes and more stable
localization, with fewer missed instances in dark regions, fewer false
positives under haze and rain, and tighter fits on small or low-contrast
objects in HDR driving scenes.

\subsection{Additional Qualitative Results for Ablation Study}
\label{app:additional_ablation}

\begin{figure*}[!t]
    \centering
    \includegraphics[width=\textwidth]{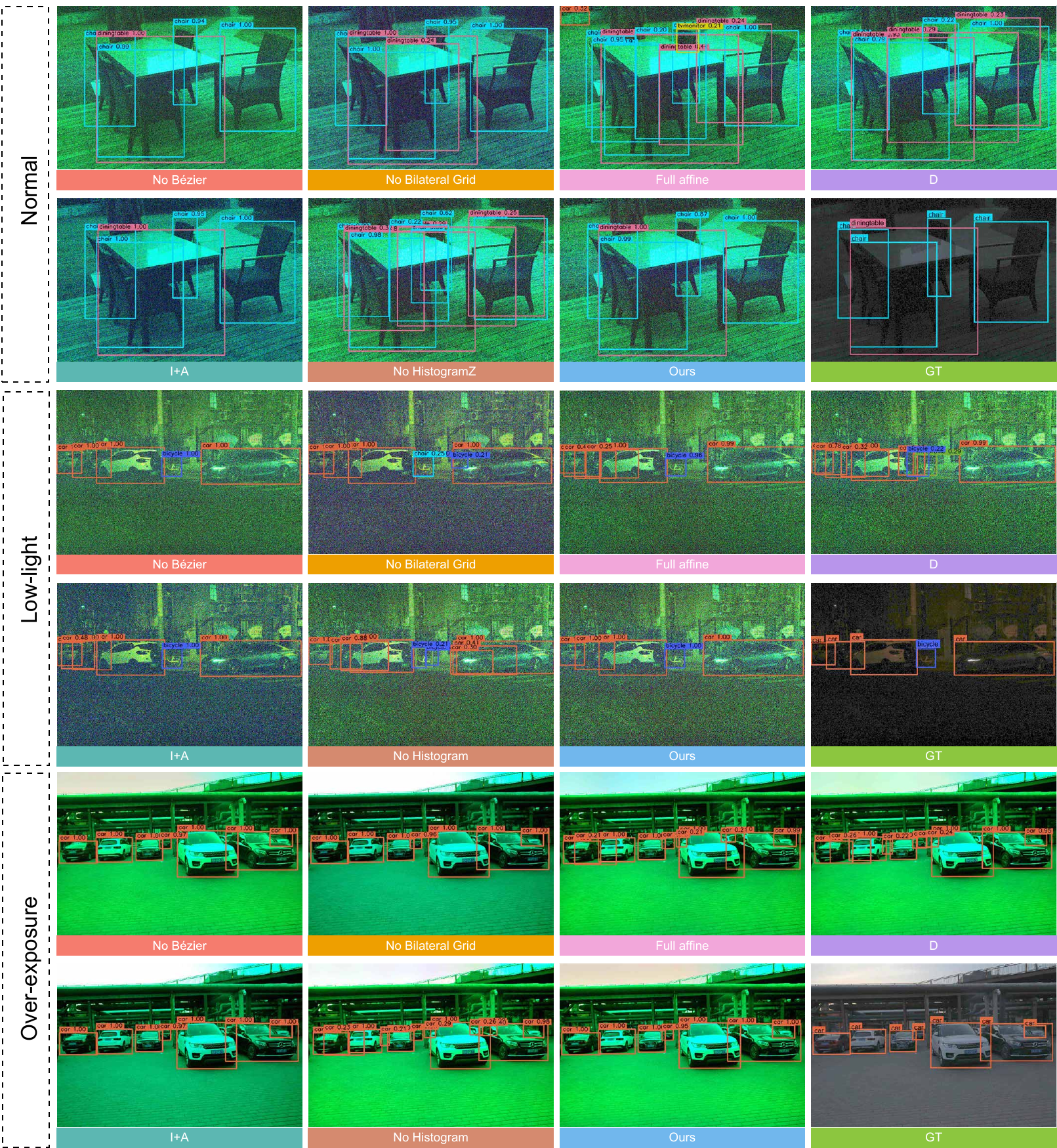}
    \caption{Qualitative ablation on representative real LOD samples under
    low-light (\textit{LOW}), normal (\textit{NM}), and over-exposure
    (\textit{OE}) conditions. Each block corresponds to one real LOD exposure
    regime and compares \emph{w/o B\'ezier}, \emph{w/o Bilateral Grid}, \emph{Affine Grid}, \emph{$\mathbf{D}$ Grid}, \emph{$\mathbf{I}{+}\mathbf{A}$ Grid}, and \emph{Ours}. Removing the global curve weakens tonal alignment, removing the bilateral grid harms spatially adaptive local refinement, and removing histogram conditioning reduces sensor-aware robustness. Our full model yields the most stable detections across all settings.}
    \label{fig:sup_ablation}
\end{figure*}

\Cref{fig:sup_ablation} visualizes the ablation variants on representative real LOD~\cite{hong2021crafting} samples spanning the three exposure regimes of low-light, normal, and over-exposure. For each regime we show the same scene processed by every ablated configuration alongside our full model and the ground-truth annotations. Removing the B\'ezier curve leaves the global tonal range poorly aligned, so detections in dark or saturated areas drift or disappear. Replacing our $\mathbf{D}(\mathbf{I}{+}\mathbf{A})$ bilateral grid with simpler \emph{Affine}, \emph{$\mathbf{D}$}, or \emph{$\mathbf{I}{+}\mathbf{A}$} variants weakens spatially adaptive local refinement and produces inconsistent boxes around small or low-contrast objects. Dropping the histogram conditioning removes sensor-aware cues, an effect most visible under over-exposure where predictions become noisier and less stable. Our full model recovers cleaner tonal structure and yields the most consistent detections across all three regimes, complementing the quantitative ablation in \Cref{sec:ablation}.

\begin{table*}[t]
\centering
\caption{Semantic segmentation on RAW ADE20K dataset~\cite{zhou2019semantic} with MiT-series backbones~\cite{xie2021segformer}, evaluated by mIoU.}
\label{tab:segmentation}
\small
\setlength{\tabcolsep}{6.2pt}
\begin{tabular}{lccccccccc}
\toprule
& \multicolumn{3}{c}{LOW} & \multicolumn{3}{c}{NM} & \multicolumn{3}{c}{OE} \\
\cmidrule(lr){2-4} \cmidrule(lr){5-7} \cmidrule(lr){8-10}
Method & B0 & B3 & B5 & B0 & B3 & B5 & B0 & B3 & B5 \\
\midrule
Linear-RAW & 0.2027 & 0.3538 & 0.3653 & 0.2632 & 0.4403 & 0.4546 & 0.2691 & 0.4229 & 0.4404 \\
\midrule
RAW-Adapter  & 0.1885 & 0.3453 & 0.3636 & 0.2939 & \thd 0.4457 & 0.4541 & 0.2683 & \thd 0.4284 & 0.4385 \\
Dr.RAW     & \sed 0.2206 & \sed 0.3581 & \sed 0.3821 & \sed 0.3134 & \sed 0.4477 & \thd 0.4666 & \sed 0.2929 & \sed 0.4343 & \thd 0.4475 \\
Dark-ISP    & \thd 0.2202 & \thd 0.3559 & \thd 0.3806 & \thd 0.3131 & 0.4432 & \sed 0.4680 & \thd 0.2917 & 0.4283 & \sed 0.4488 \\
\textbf{Ours} & \fst \textbf{0.2872} & \fst \textbf{0.3957} & \fst \textbf{0.4082} & \fst \textbf{0.3534} & \fst \textbf{0.4516} & \fst \textbf{0.4708} & \fst \textbf{0.3372} & \fst \textbf{0.4381} & \fst \textbf{0.4560} \\
\bottomrule
\end{tabular}
\end{table*}

\subsection{Additional Qualitative Results for RAW Semantic Segmentation}
\label{app:additional_segmentation}

\paragraph{Implementation Details.}
We follow the exposure-curated ADE20K~\cite{zhou2019semantic} protocol of RAW-Adapter ~\cite{cui2024raw}, with Low-light (\textit{LOW}), Normal (\textit{NM}), and Over-exposure (\textit{OE}) splits. The network input is an offline-processed 8-bit RGB image derived from RAW data and normalized to $[0,1]$, following the released protocol. We insert the same histogram-conditioned global-local adapter described in \Cref{sec:bezier,sec:grid,sec:arch} into SegFormer with MiT-B0, MiT-B3, and MiT-B5 backbones~\cite{xie2021segformer} and keep the decoder unchanged. We compare with Linear-RAW, RAW-Adapter ~\cite{cui2024raw}, Dr.RAW~\cite{huang2025dr}, and Dark-ISP~\cite{guo2025dark}, the baselines with compatible segmentation implementations under the same MMSegmentation framework.

Training runs for $200\mathrm{k}$ iterations using AdamW (base learning rate $6\times10^{-5}$, $\beta_1{=}0.9$, $\beta_2{=}0.999$, weight decay $0.01$), with a $1.5\mathrm{k}$-iteration linear warmup followed by polynomial decay. We use a global batch size of $16$ on a single H100 GPU and apply a $10\times$ learning-rate multiplier to the adapter and decode head. MiT-B0 and MiT-B5 are initialized from the corresponding ImageNet-pretrained SegFormer checkpoints, while MiT-B3 is warm-started from an earlier $120\mathrm{k}$ checkpoint. Data augmentation uses random resize to $(2048,512)$ with ratio range $[0.5,2.0]$, random crop to $512\times512$, and random horizontal flipping. We report mean Intersection-over-Union (mIoU) with single-scale sliding-window inference and no test-time augmentation.

\paragraph{Quantitative Results.}
\Cref{tab:segmentation} reports mIoU across the three exposure splits and three backbone scales. Our method consistently improves over all baselines in every setting. The gain is largest with the lightweight MiT-B0 backbone, where we surpass the strongest baseline by $6.66$, $4.00$, and $4.43$ mIoU points under \textit{LOW}, \textit{NM}, and \textit{OE}, respectively. The global B\'ezier curve stabilizes image-level tonal and exposure statistics, while the bilateral grid refines boundary-sensitive local color and contrast, so the two
components together improve dense prediction without modifying the segmentation decoder.

\begin{figure}[!tp]
    \centering
    \includegraphics[width=\textwidth]{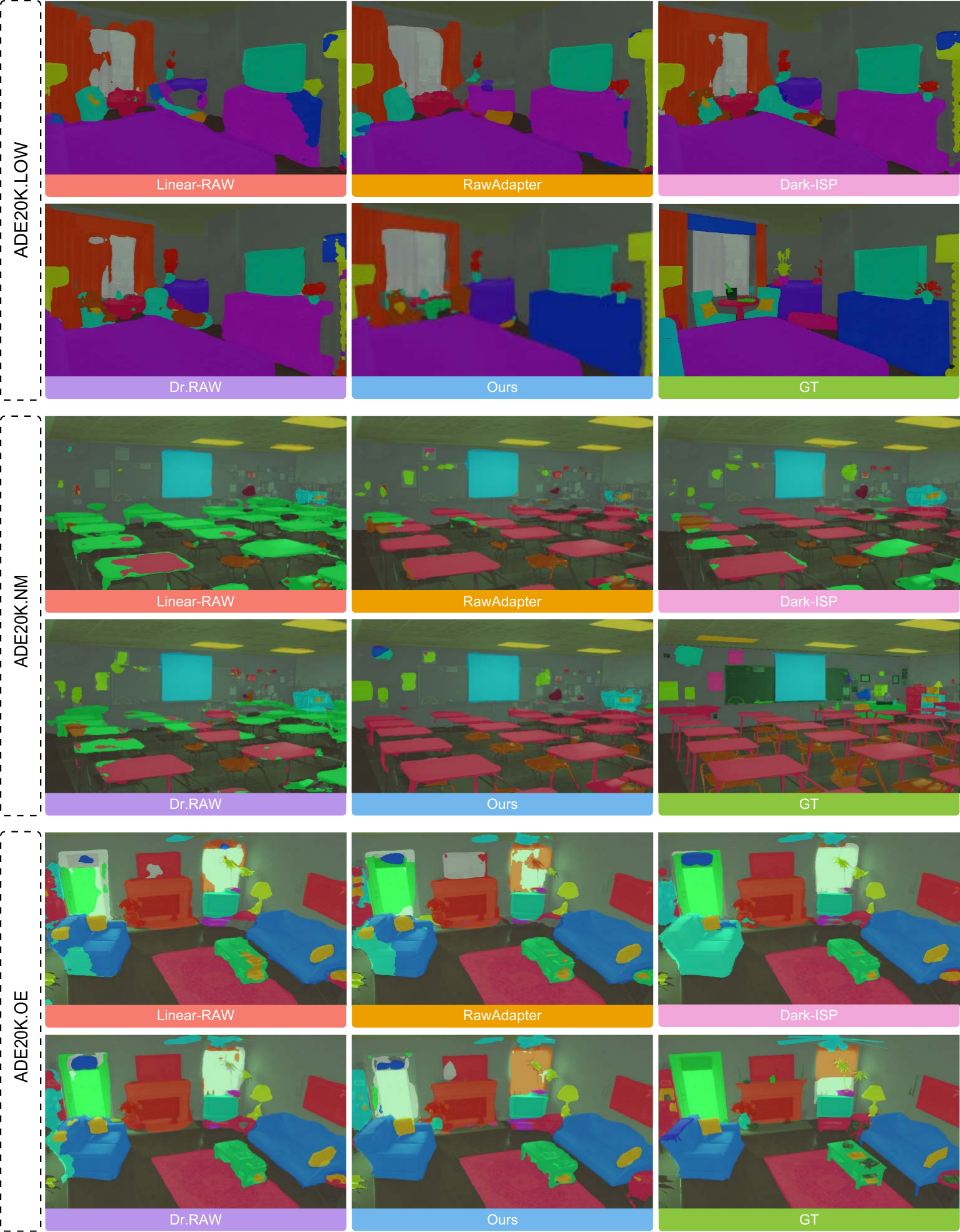}
    \caption{Qualitative transfer results on exposure-curated ADE20K under three illumination regimes: Low-light (\textit{LOW}), Normal (\textit{NM}), and Over-exposure (\textit{OE}). Our method yields more coherent regions and cleaner semantic boundaries across all three settings.}
    \label{fig:sup_segmentation}
\end{figure}

\paragraph{Qualitative Results.}
\Cref{fig:sup_segmentation} shows qualitative comparisons. Under low-light inputs, baseline methods often produce fragmented semantic regions or lose small objects in dark areas. Under over-exposed inputs, saturated regions and compressed local contrast make object boundaries
difficult to recover. Our method produces more coherent semantic regions and better preserves object boundaries across \textit{LOW}, \textit{NM}, and \textit{OE}, indicating that the same
histogram-conditioned global-local adapter transfers from box prediction to dense prediction under RAW exposure variation.

\clearpage

\bibliographystyle{plainnat}
\bibliography{sample}

\end{document}